\documentclass{article}

\usepackage{tikz}
\tikzset{
  pics/carc/.style args={#1:#2:#3}{
    code={
      \draw[pic actions] (#1:#3) arc(#1:#2:#3);
    }
  }
}

\usepackage[utf8]{inputenc} % allow utf-8 input
\usepackage[T1]{fontenc}    % use 8-bit T1 fonts
\usepackage{hyperref}       % hyperlinks
\usepackage{url}            % simple URL typesetting
\usepackage{booktabs}       % professional-quality tables
\usepackage{amsfonts}       % blackboard math symbols
\usepackage{nicefrac}       % compact symbols for 1/2, etc.
\usepackage{microtype}      % microtypography
\usepackage{wrapfig,lipsum}
\usepackage{comment}

\usepackage{graphicx}

\usepackage{bm}
\usepackage[export]{adjustbox}
\usepackage{amsmath}
\usepackage{amsthm}
\usepackage{amsbsy}
\usepackage{calc}
\usepackage{color}
\usepackage{float}
\usepackage{multirow}
\usepackage{todonotes}

\usepackage{amsmath}

\usepackage{subcaption}

\usepackage[accepted]{icml_arxiv}

\newcommand\corobustness{\mathbb{P}_{x\sim q}[x \notin E]}
\newcommand\advrobustness{\mathbb{P}_{x\sim p}[d(x,E) > \epsilon]}

\newcommand\errorvolume{\mathbb{P}_{x\sim q}[x \in E]}
\newcommand\errorboundary{\mathbb{P}_{x\sim p}[d(x,E)\le\epsilon]}
\newcommand\errorboundarynoisy{\mathbb{P}_{x\sim q}[d(x,E)\le\epsilon]}

\newtheorem*{thm}{Theorem}

\begin{document}

\twocolumn[
\icmltitle{Adversarial Examples Are a Natural Consequence of Test Error in Noise}

% It is OKAY to include author information, even for blind
% submissions: the style file will automatically remove it for you
% unless you've provided the [accepted] option to the icml2019
% package.

% List of affiliations: The first argument should be a (short)
% identifier you will use later to specify author affiliations
% Academic affiliations should list Department, University, City, Region, Country
% Industry affiliations should list Company, City, Region, Country

% You can specify symbols, otherwise they are numbered in order.
% Ideally, you should not use this facility. Affiliations will be numbered
% in order of appearance and this is the preferred way.
\icmlsetsymbol{equal}{*}

\begin{icmlauthorlist}
\icmlauthor{Nicolas Ford}{equal,brain,aires}
\icmlauthor{Justin Gilmer}{equal,brain}
\icmlauthor{Nicholas Carlini}{brain}
\icmlauthor{Ekin D. Cubuk}{brain}
\end{icmlauthorlist}

\icmlaffiliation{brain}{Google Brain}
\icmlaffiliation{aires}{This work was completed as part of the Google AI Residency}

\icmlcorrespondingauthor{Nicolas Ford}{nicf@google.com}
\icmlcorrespondingauthor{Justin Gilmer}{gilmer@google.com}

% You may provide any keywords that you
% find helpful for describing your paper; these are used to populate
% the "keywords" metadata in the PDF but will not be shown in the document
\icmlkeywords{Machine Learning, ICML}

\vskip 0.3in
]

\printAffiliationsAndNotice{\icmlEqualContribution}

\begin{abstract}
    Over the last few years, the phenomenon of \emph{adversarial examples} --- maliciously constructed inputs that fool trained machine learning models --- has captured the attention of the research community, especially when the adversary is restricted to small modifications of a correctly handled input. Less surprisingly, image classifiers also lack human-level performance on randomly corrupted images, such as images with additive Gaussian noise. In this paper we provide both empirical and theoretical evidence that these are two manifestations of the same underlying phenomenon, establishing close connections between the adversarial robustness and corruption robustness research programs. This suggests that improving adversarial robustness should go hand in hand with improving performance in the presence of more general and realistic image corruptions. Based on our results we recommend that future adversarial defenses consider evaluating the robustness of their methods to distributional shift with benchmarks such as Imagenet-C.
    %We establish this connection in several ways. First, we find that adversarial examples exist at the same distance scales we would expect from a linear model with the same performance on corrupted images. Next, we show that Gaussian data augmentation during training improves robustness to small adversarial perturbations and that adversarial training improves robustness to several types of image corruptions. Finally, we present a model-independent upper bound on the distance from a \emph{corrupted} image to its nearest error given test performance and show that in practice we already come close to achieving the bound, so that improving robustness further for the corrupted image distribution requires significantly reducing test error.
 % withe common corruption benchmark as a computationally tractable evaluation metric for adversarial defenses to consider.
\end{abstract}

\section{Introduction}
    \label{sec:introduction}
    State-of-the-art computer vision models can achieve impressive performance on many image classification tasks. Despite this, these same models still lack the robustness of the human visual system to various forms of image corruptions. For example, they are distinctly subhuman when classifying images distorted with additive Gaussian noise \citep{dodge2017study}, they lack robustness to different types of blur, pixelation, and changes in brightness \citep{hendrycks2018benchmarking}, lack robustness to random translations of the input \citep{azulay2018deep}, and even make errors when foreign objects are inserted into the field of view \citep{rosenfeld2018elephant}. At the same time, they are also sensitive to small, worst-case perturbations of the input, so-called ``adversarial examples’’ \citep{Szegedy14}. This latter phenomenon has struck many in the machine learning community as surprising and has attracted a great deal of research interest, while the former has received considerably less attention.

    The machine learning community has researchers working on each of these two types of errors: adversarial example researchers seek to measure and improve robustness to small-worst case perturbations of the input while corruption robustness researchers seek to measure and improve model robustness to distributional shift. In this work we analyze the connection between these two research directions, and we see that adversarial robustness is closely related to robustness to certain kinds of distributional shift. In other words, the existence of adversarial examples follows naturally from the fact that our models have nonzero test error in certain corrupted image distributions.
    
    We make this connection in several ways. First, in Section~\ref{sec:cleanimage}, we provide a novel analysis of the error set of an image classifier. We see that, given the error rates we observe in Gaussian noise, the small adversarial perturbations we observe in practice appear at roughly the distances we would expect from a \emph{linear} model, and that therefore there is no need to invoke any strange properties of the decision boundary to explain them. This relationship was also explored in  \citet{fawzi2018empirical, fawzi2016robustness}.
    
    In Section~\ref{sec:isoperimetric}, we show that improving an alternate notion of adversarial robustness \emph{requires} that error rates under large additive noise be reduced to essentially zero.
    
    Finally, this connection suggests that methods which are designed to increase the distance to the decision boundary should also improve robustness to Gaussian noise, and vice versa. In Section~\ref{sec:advtrain} we confirm that this is true by examining both adversarially trained models and models trained with additive Gaussian noise. We also show that measuring corruption robustness can effectively distinguish successful adversarial defense methods from ones that merely cause vanishing gradients.
    
    We hope that this work will encourage both the adversarial and corruption robustness communities to work more closely together, since their goals seem to be so closely related. In particular, it is not common for adversarial defense methods to measure corruption robustness. Given that successful adversarial defense methods should also improve some types of corruption robustness we recommend that future researchers consider evaluating corruption robustness in addition to adversarial robustness.

\section{Related Work}
    The broader field of \emph{adversarial machine learning} studies general ways in which an adversary may interact with an ML system, and dates back to 2004 \citep{dalvi2004adversarial, biggio2018wild}. Since the work of \citet{Szegedy14}, a subfield has focused specifically on the phenomenon of small adversarial perturbations of the input, or ``adversarial examples.'' Many algorithms have been developed to find the smallest perturbation in input space which fool a classifier \citep{carlini2017adversarial, madry2017advexamples}. Defenses have been proposed for increasing the robustness of classifiers to small adversarial perturbations, however many have later been shown ineffective \citep{carlini2017adversarial}. To our knowledge the only method which has been confirmed by a third party to increase $l_p$-robustness (for certain values of $\epsilon$) is adversarial training \citep{madry2017advexamples}. However, this method remains sensitive to slightly larger perturbations \citep{chen2017madry}. % In \citet{Szegedy14} it was proposed these adversarial examples occupy a dense, measure-zero subset of image space. However, more recent work has provided evidence that this is not true. For example, \citet{fawzi2016robustness, franceschi2018robustness} shows that under linearity assumptions of the decision boundary small adversarial perturbations exist when test error in additive noise is non-zero. 
    
    Several recent papers \citep{gilmer2018adversarial,mahloujifar2018curse, dohmatob2018limitations, fawzi2018adversarial} use \emph{concentation of measure} to prove rigorous upper bounds on adversarial robustness for certain distributions in terms of test error, suggesting non-zero test error may imply the existence of adversarial perturbations. This may seem in contradiction with empirical observations that increasing small perturbation robustness tends to reduce model accuracy \citep{tsipras2018robustness}. We note that these two conclusions are not necessarily in contradiction to each other. It could be the case that hard bounds on adversarial robustness in terms of test error exist, but current classifiers have yet to approach these hard bounds.  %Furthermore, this tradeoff need not hold absolutely \citep{stutz2018disentangling}. %It is possible that concentration of measure explains why adversarially trained models remain sensitive to perturbations at larger $\epsilon$. %In fact, such bounds hold for the multivariate Gaussian distribution \cite{mahloujifar2018curse}. In particular, any model 
    
    Because we establish a connection between adversarial robustness and model accuracy in \emph{corrupted} image distributions, our results do not contradict reports that adversarial training reduces accuracy in the \emph{clean} distribution \citep{tsipras2018robustness}. In fact, we find that improving adversarial robustness also improves corruption robustness. %In fact, we observe that several methods which improve accuracy in Gaussian noise also degrade clean test performance. %In fact, we observe that improving the accuracy in corrupted image distributions also trades off with clean test accuracy. %Towards the other direction, Gaussian data augmentation has been explored as an efficient method for increasing adversarial robustness \cite{zantedeschi2017efficient}. In this work we also explore the effects of Gaussian data augmentation on adversarial robustness. %This observation is predicted by the ``locally flat decision boundary model'' first introduced in \cite{fawzi2016robustness, franceschi2018robustness}, (discussed further in Section~\ref{sec:cleanimage}) 
    
    %\todo{"efficient defenses against adversarial attack" aisec'17 proposes training on gaussian noise}

\section{Adversarial and Corruption Robustness}
    \label{sec:math}
    Both adversarial robustness and corruption robustness can be thought of as functions of the \textbf{error set} of a statistical classifier. This set, which we will denote $E$, is the set of points in the input space on which the classifier makes an incorrect prediction. In this paper we will only consider perturbed versions of training or test points, and we will always assume the input is corrupted such that the ``correct'' label for the corrupted point is the same as for the clean point. This assumption is commonly made in works which study model robustness to random corruptions of the input \citep{hendrycks2018benchmarking, dodge2017study}.
    
    Because we are interested in how our models perform on both clean images and corrupted ones, we introduce some notation for both distributions. We will write $p$ for the \textit{natural} image distribution, that is, the distribution from which the training data was sampled. We will use $q$ to denote whichever \textit{corrupted} image distribution we are working with. A sample from $q$ will always look like a sample from $p$ with a random corruption applied to it, like some amount of Gaussian noise. Some examples of noisy images can be found in Figure~\ref{fig:optimal_vis} in the appendix.
    
    We will be interested in two quantities. The first, \textbf{corruption robustness} under a given corrupted image distribution $q$, is $\corobustness$, the probability that a random sample from the $q$ is not an error. The second is called \textbf{adversarial robustness}. For a clean input $x$ and a metric on the input space $d$, let $d(x,E)$ denote the distance from $x$ to the nearest point in $E$. The adversarial robustness of the model is then $\advrobustness$, the probability that a random sample from $p$ is not within distance $\epsilon$ of some point in the error set. When we refer to ``adversarial examples'' in this paper, we will always mean these nearby errors.
    
        %\label{sec:models}
    In this work we will investigate several different models trained on the CIFAR-10 and ImageNet datasets. For CIFAR-10 we look at the naturally trained and adversarially trained models which have been open-sourced by \citet{madry2017advexamples}. We also trained the same model on CIFAR-10 with Gaussian data augmentation. For ImageNet, we investigate an Inception v3~\cite{szegedy16} trained with Gaussian data augmentation. In all cases, Gaussian data augmentation was performed by first sampling a $\sigma$ uniformly between 0 and some specified upper bound and then adding random Gaussian noise at that scale. Additional training details can be found in Appendix~\ref{app:trainingdetails}. We were unable to study the effects of adversarial training on ImageNet because no robust open sourced model exists. (The models released in \citet{tramer2017ensemble} only minimally improve robustness to the white box PGD adversaries we consider here.)

\section{Errors in Gaussian Noise Suggest Adversarial Examples}
    \label{sec:cleanimage}
    
    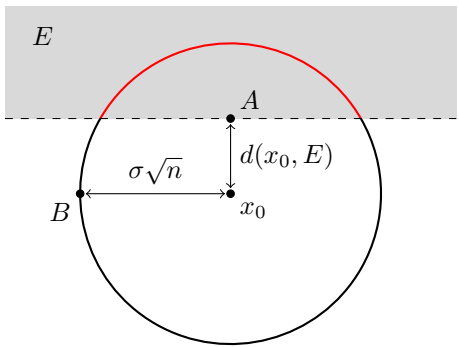
\begin{figure}[t]
     \centering
     \begin{tikzpicture}

        \fill[gray!30] (-3,1) rectangle (3,2.5);
     
        \draw[black, thick] (0,0) pic{carc=150:390:2};
        \draw[red, thick] (0,0) pic{carc=30:150:2};
        \draw[black, dashed] (-3,1) -- (3,1);
        
        \draw[fill] (0,0) circle [radius=0.05];
        \node[below right] at (0,0) {$x_0$};
        \node at (-2.5,2.1) {$E$};
        
        \draw[fill] (0,1) circle [radius=0.05];
        \node[above right] at (0,1) {$A$};
        
        \draw[black, <->] (180:1.925) -- (180:0.075) node[pos=0.5, above]{$\sigma\sqrt{n}$};
        \draw[fill] (180:2) circle [radius=0.05];
        \node[below left] at (180:2) {$B$};
        \draw[black, <->] (0,0.075) -- (0,0.925) node[pos=0.5, right]{$d(x_0,E)$};

     \end{tikzpicture}

     \caption{\label{fig:halfspace} When the input dimension, $n$, is large and the model is linear, even a small error rate in additive noise implies the existence of small adversarial perturbations. For a point $x_0$ in image space, most samples from $\mathcal{N}(x_0;\sigma^2I)$ (point $B$) lie close to a sphere of radius $\sigma\sqrt{n}$ around $x_0$, drawn here as a circle. For a linear model the error set $E$ is a half-space, and the error rate $\mu$ is approximately equal to the fraction of the sphere lying in this half-space. The distance $d(x_0,E)$ from $x_0$ to its nearest error (point $A$) is also drawn. Note the relationship between $\sigma$, $\mu$, and $d(x_0,E)$ does not depend on the dimension. However, because the typical distance to a sample from the Gaussian is $\sigma\sqrt{n}$ the ratio between the distance from $x_0$ to $A$ and the distance from $x_0$ to $B$ shrinks as the dimension increases.}
    \end{figure}

    \begin{figure*}[t]
      \centering
      \begin{subfigure}{.45\textwidth}
        \includegraphics[width=1.0\linewidth]{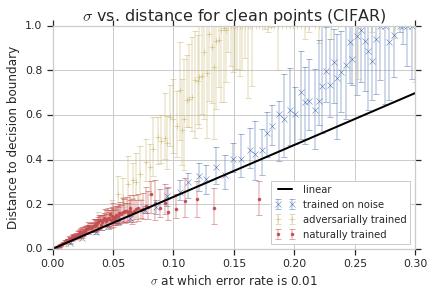}
      \end{subfigure}
      \hspace{.05\textwidth}
      \begin{subfigure}{.45\textwidth}
        \includegraphics[width=1.0\linewidth]{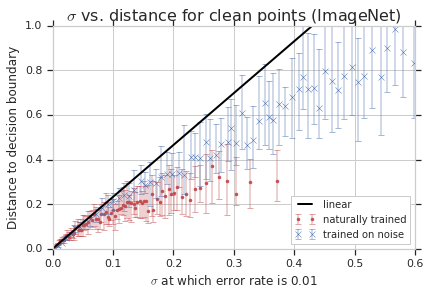}
      \end{subfigure}
      \begin{subfigure}{.45\textwidth}
        \includegraphics[width=1.0\linewidth]{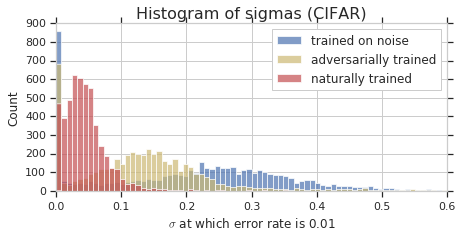}
      \end{subfigure}
      \hspace{.05\textwidth}
      \begin{subfigure}{.45\textwidth}
        \includegraphics[width=1.0\linewidth]{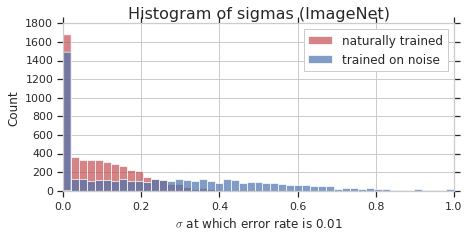}
      \end{subfigure}
      \caption{\label{fig:distancevsvolume}\textit{(Top)} Comparing the $l_2$ distance to the decision boundary with the $\sigma$ for which the error rate in Gaussian noise is 1\%. Each point represents 50 images from the test set, and the median values for each coordinate are shown. The error bars cover the 25th to 75th percentiles. The PGD attack was run with $\epsilon=1$, so the distances to the decision boundary reported here are cut off at 1. \textit{(Bottom)} Histograms of the $x$ coordinates from the above plots. A misclassified point is assigned $\sigma=0$.}
    \end{figure*}
    
    \begin{figure*}[t]
     \centering
     \begin{subfigure}{.4\textwidth}
     \includegraphics[width=1.0\linewidth]{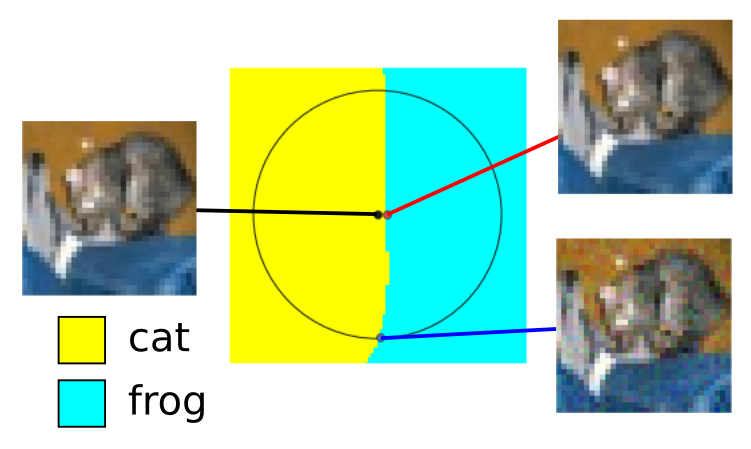}
     \end{subfigure}
     \hspace{0.05\textwidth}
     \begin{subfigure}{.4\textwidth}
     \includegraphics[width=1.0\linewidth]{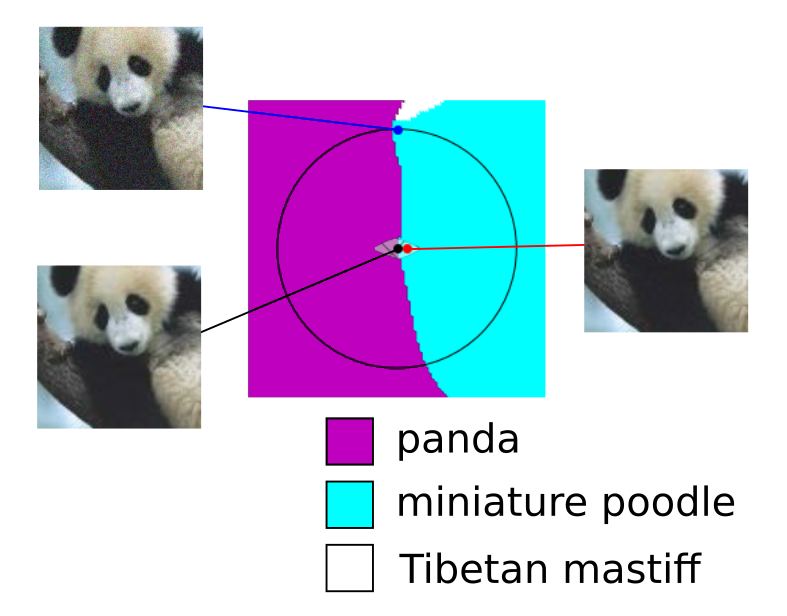}
     \end{subfigure}
     %\hspace{0.05\textwidth}
     %\includegraphics[width=0.45\linewidth]{/weird_slice.png}
     %\hspace{.05\textwidth}
     %\centering
     %  \begin{subfigure}{.55\textwidth}
     %  \includegraphics[width=1.0\linewidth]{very_adversarial_pandas.png}
     %\end{subfigure}
     \caption{\label{fig:pandas} Two-dimensional slices of image space together with the classes assigned by trained models. Each slice goes through three points, a clean image from the test set (black), an error found by randomly perturbing the center image with Gaussian noise (blue), and an error found using a targeted PGD attack (red). The black circles have radius $\sigma \sqrt{n}$, indicating the typical size of the Gaussian perturbation used. The diamond-shaped region in the center of the right image shows the $l_\infty$ ball of radius $8/255$. In both slices, the decision boundary resembles a half-space as predicted in Figure~\ref{fig:halfspace}, demonstrating how non-zero error rate in noise predicts the existence of small adversarial perturbations. The CIFAR-10 model on the left was evaluated with $\sigma=0.04$ (black circle has radius $2.22$), where $0.21$\% of Gaussian perturbations are classified as ``frog'' (cyan region). The adversarial error was found at distance $0.159$ while the half-space model predicts errors at distance $0.081$. The ImageNet model on the right was evaluated at $\sigma=0.08$ (black circle has radius $31.4$) where $0.1$\% of Gaussian perturbations were misclassified as ``miniture poodle'' (cyan). The adversarial error has distance $0.189$ while the half-space model predicts errors at distance $0.246$. For the panda picture on the right we also found closer errors than what is shown by using an untargeted attack (an image was assigned class ``indri'' at distance $0.024$). Slices showing more complicated behavior can be found in Appendix~\ref{app:church}.} %Slices which show more complicated decision boundaries can be found in the Appendix.} % In particular we found an image assigned class ``indri'' at distance $0.024$,   The black circle has radius $31.4$, corresponding to noise with $\sigma=31.4/\sqrt{n}=0.08$. We see an image from the test set (black), a random misclassified Gaussian perturbation at standard deviation $0.08$ (blue), and an error found using a targeted PGD attack (red). The estimated measure of the cyan region (``miniature poodle'') in the Gaussian distribution is about 0.1\%. A linear model with this same error rate on noise at this scale would have an error at distance $0.246$. A PGD attack targeted at the class ``miniature poodle'' finds an error at distance $0.189$ (this is the error shown).  but this point in fact has another error, assigned the class ``indri'' at distance $0.024$. The small diamond-shaped region in the center of the image is the $l_\infty$ ball of radius 8/255.}
     %\textit{(Right)} A slice at a larger scale with the same black point, together with an error from the clean set (blue) and an adversarially constructed error (red) which are both assigned to the same class (``elephant'').}
    \end{figure*}
    
    We will start by examining the relationship between adversarial and corruption robustness in the case where $q$ consists of images with additive Gaussian noise.
    
    \textbf{The Linear Case.} For linear models, the error rate in Gaussian noise exactly determines the distance to the decision boundary. This observation was also made in \citet{fawzi2016robustness, fawzi2018empirical}. % relationship between errors in Gaussian noise and $l_2$ adversarial examples is exact.
    %\todo{what does "exact" mean? Perfectly correlated?}
    
    It will be useful to keep the following intuitive picture in mind. In high dimensions, most samples from the Gaussian distribution $\mathcal{N}(x_0;\sigma^2I)$ lie close to the surface of a sphere of radius $\sigma$ centered at $x_0$. The decision boundary of a linear model is a plane, and since we are assuming that the ``correct'' label for each noisy point is the same as the label for $x_0$, our error set is simply the half-space on the far side of this plane.
    
    The relationship between adversarial and corruption robustness corresponds to a simple geometric picture. If we slice a sphere with a plane, as in Figure~\ref{fig:halfspace}, the distance to the nearest error is equal to the \emph{distance} from the plane to the center of the sphere, and the corruption robustness is the fraction of the \emph{surface area} cut off by the plane. This relationship changes drastically as the dimension increases: most of the surface area of a high-dimensional sphere lies very close to the equator, which means that cutting off even, say, 1\% of the surface area requires a plane which is very close to the center. Thus, for a linear model, even a relatively small error rate on Gaussian noise implies the existence of errors very close to the clean image (i.e., an adversarial example).
    
    To formalize this relationship, pick some clean image $x_0$ and consider the Gaussian distribution $\mathcal{N}(x_0;\sigma^2I)$. For a fixed $\mu$, let $\sigma(x_0,\mu)$ be the $\sigma$ for which the error rate is $\mu$, that is, for which \[\mathbb{E}_{x\sim\mathcal{N}(x_0;\sigma^2I)}[x\in E]=\mu.\] Then, letting $d$ denote $l_2$ distance, we have \begin{equation}
        \label{eqn:distscale}
        d(x_0,E)=-\sigma(x_0,\mu)\Phi^{-1}(\mu),
    \end{equation}
    where \[\Phi(t)=\frac{1}{\sqrt{2\pi}}\int_{-\infty}^t\exp(-x^2/2)dx\] is the cdf of the univariate standard normal distribution. (Note that $\Phi^{-1}(\mu)$ is negative when $\mu<\frac12$.)
    
    This expression depends only on the error rate $\mu$ and the standard deviation $\sigma$ of a single component, and not directly on the dimension, but the dimension appears if we consider the distance from $x_0$ to a typical sample from $\mathcal{N}(x_0;\sigma^2I)$, which is $\sigma\sqrt{n}$. When the dimension is large the distance to the decision boundary will be significantly smaller than the distance to a noisy image.
    
    For example, this formula says that a linear model with an error rate of $0.01$ in noise with $\sigma=0.1$ will have an error at distance about $0.23$. In three dimensions, a typical sample from this noise distribution will be at a distance of around $0.1\sqrt{3}\approx 0.17$. However when $n=3072$, the dimension of the CIFAR-10 image space, these samples lie at a distance of about $5.54$. So, in the latter case, a 1\% error rate on random perturbations of size $5.54$ implies an error at distance $0.23$, more than 20 times closer. Detailed curves showing this relationship can be found in Appendix~\ref{app:curve_vis}.
    
    \textbf{Comparing Neural Networks to the Linear Case.} The decision boundary of a neural network is, of course, not linear. However, by comparing the ratio between $d(x_0,E)$ and $\sigma(x_0,\mu)$ for neural networks to what it would be for a linear model, we can investigate the relationship between adversarial and corruption robustness. We ran experiments on several neural network image classifiers and found results that closely resemble Equation~\ref{eqn:distscale}. Adversarial examples therefore are not ``surprisingly'' close to $x_0$ given the performance of each model in Gaussian noise. 
    %\todo{I think goodfellow et al 2015 also makes this argument}
    
    Concretely, we examine this relationship when $\mu=0.01$. For each test point, we compare $\sigma(x_0,0.01)$ to an estimate of $d(x_0,E)$. Because it is not feasible to compute $d(x_0,E)$ exactly, we instead search for an error using PGD \citep{madry2017advexamples} and report the nearest error we can find.
    
    Figure~\ref{fig:distancevsvolume} shows the results for several CIFAR-10 and ImageNet models, including ordinarily trained models, models trained with Gaussian data augmentation with $\sigma=0.4$, and an adversarially trained CIFAR-10 model. We also included a line representing how these quantities would be related for a linear model, as in Equation~\ref{eqn:distscale}. Because most test points lie close to the predicted relationship for a linear model, we see that the half-space model shown in Figure~\ref{fig:halfspace} accurately predicts the existence of small perturbation adversarial examples.
    
    It is interesting to observe how each training procedure affected the two quantities we measured. First, adversarial training and Gaussian data augmentation increased \emph{both} $\sigma(x_0,0.01)$ and $d(x_0,E)$ on average. The adversarially trained model deviates from the linear case the most, but it does so in the direction of \emph{greater} distances to the decision boundary. While both augmentation methods do improve both quantities, Gaussian data augmentation had a greater effect on $\sigma$ (as seen in the histograms) while adversarial training had a greater effect on $d$. We explore this further in Section~\ref{sec:advtrain}.
    
    \textbf{Visual Confirmation of the Half-space Model} In Figure~\ref{fig:pandas} we draw two-dimensional slices in image space through three points. (Similar visualizations have appeared in \citet{fawzi2018empirical}, and are called ``church window plots.'')
    
    This visualized decision boundary closely matches the half-space model in Figure~\ref{fig:halfspace}. We see that an error found in Gaussian noise lies in the same connected component of the error set as an error found using PGD, and that at this scale that component visually resembles a half-space. This figure also illustrates the connection between adversarial example research and corruption robustness research. To measure adversarial robustness is to ask whether or not there are any errors in the $l_{\infty}$ ball --- the small diamond-shaped region in the center of the image --- and to measure corruption robustness is to measure the volume of the error set in the defined noise distribution. At least in this slice, nothing distinguishes the PGD error from any other point in the error set apart from its proximity to the clean image.
    
    We give many more church window plots in Appendix~\ref{app:church}.
    
    %The figure on the right shows a different slice through the same test point but at a larger scale. This slice includes an ordinary test error along with an adversarial perturbation of the center image constructed with the goal of maintaining visual similarity while having a large $l_2$ distance. The two errors are both classified (incorrectly) by the model as ``elephant.'' This adversarial error is actually \emph{farther} from the center than the test error, but they still clearly belong to the same connected component. This suggests that defending against worst-case content-preserving perturbations \citep{gilmer2018motivating} requires removing all errors at a scale comparable to the distance between unrelated pairs of images.

\section{Concentration of Measure for Noisy Images}
    \label{sec:isoperimetric}
    There is an existing research program \citep{gilmer2018adversarial,mahloujifar2018curse, dohmatob2018limitations} which proves hard upper bounds on adversarial robustness in terms of the error rate of a model. This phenomenon is sometimes called \emph{concentration of measure}. Because proving a theorem like this requires understanding the distribution in question precisely, these results typically deal with simple ``toy'' distributions rather than those corresponding to real data. In this section we take a first step toward bridging this gap. By comparing our models to a classical concentration of measure bound for the Gaussian distribution, we gain another perspective on our motivating question.
    
    \textbf{The Gaussian Isoperimetric Inequality.} As in Section~\ref{sec:cleanimage}, let $x_0$ be a correctly classified image and consider the distribution $q=\mathcal{N}(x_0;\sigma^2I)$. Note $q$ is the distribution of random Gaussian perturbations of $x_0$. The previous section discussed the distance from $x_0$ to its nearest error. In this section we will instead discuss the distance from a typical sample from $q$ (e.g. point $B$ in Figure~\ref{fig:halfspace}) to its nearest error. %(In Section~\ref{sec:cleanimage}, we were concerned with the distance from $x_0$ to its nearest error. In this section, we will instead investigate the distance from a typical sample from $q$ to its nearest error.\todo{I don't understand the difference. is it in the fact that $q$ is random?}
    
    For random samples from $q$, there is a precise sense in which small adversarial perturbations exist only because test error is nonzero. That is, given the error rates we actually observe on noisy images, most noisy images \emph{must} be close to the error set. This result holds completely independently of any assumptions about the model and follows from a fundamental geometric property of the Gaussian distribution, which we will now make precise.
    
    \begin{figure*}[ht]
     \centering
     \begin{subfigure}{\linewidth}
        \centering
        \includegraphics[width=0.45\linewidth]{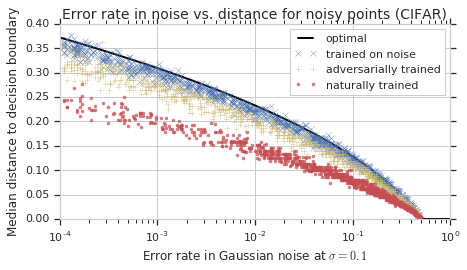}
        \hspace{0.05\linewidth}
        \includegraphics[width=0.45\linewidth]{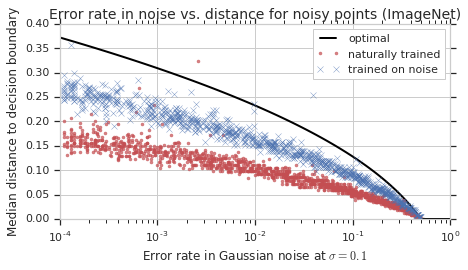}
     \end{subfigure}
     \vspace{1ex}
     \begin{subfigure}{\linewidth}
        \centering
        \includegraphics[width=0.45\linewidth]{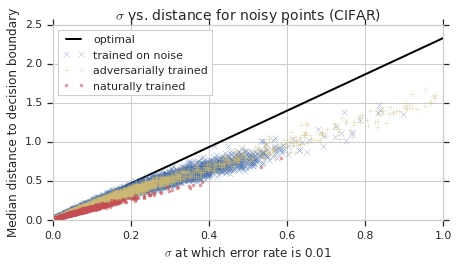}
        \hspace{0.05\linewidth}
        \includegraphics[width=0.45\linewidth]{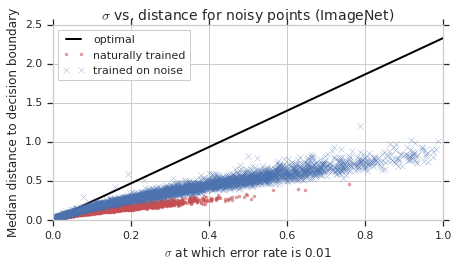}
     \end{subfigure}
     \caption{\label{fig:isoplot}These plots give two ways to visualize the relationship between the error rate in noise and the distance from noisy points to the decision boundary (found using PGD). Each point on each plot represents one image from the test set. On the top row, we compare the error rate of the model with Gaussian perturbations at $\sigma=0.1$ to the distance from the median \emph{noisy} point to its nearest error. On the bottom row, we compare the $\sigma$ at which the error rate is $0.01$ to this same median distance. (These are therefore similar to the plots in Figure~\ref{fig:distancevsvolume}.) The thick black line at the top of each plot is the upper bound provided by the Gaussian isoperimetric inequality. We include data from a model trained on clean images, an adversarially trained model, and a model trained on Gaussian noise ($\sigma=0.4$.)}
    \end{figure*}

    Let $\epsilon^*_q(E)$ be the median distance from one of these noisy images to the nearest error. (In other words, it is the $\epsilon$ for which $\errorboundarynoisy=\frac12$.) As before, let $\errorvolume$ be the probability that a random Gaussian perturbation of $x_0$ lies in $E$. It is possible to deduce a bound relating these two quantities from the \emph{Gaussian isoperimetric inequality} \citep{borell1975brunn}. The form we will use is:
    
    \begin{thm}[Gaussian Isoperimetric Inequality]
        Let $q=\mathcal{N}(0;\sigma^2I)$ be the Gaussian distribution on $\mathbb{R}^n$ with variance $\sigma^2I$, and, for some set $E\subseteq\mathbb{R}^n$, let $\mu=\errorvolume$.
    
        As before, write $\Phi$ for the cdf of the univariate standard normal distribution. If $\mu\ge\frac12$, then $\epsilon^*_q(E)=0$. Otherwise, $\epsilon^*_q(E)\le -\sigma\Phi^{-1}(\mu)$, with equality when $E$ is a half space.
    \end{thm}
    
    In particular, for any machine learning model for which the error rate in the distribution $q$ is at least $\mu$, the median distance to the nearest error is at most $-\sigma\Phi^{-1}(\mu)$. Because each coordinate of a multivariate normal is a univariate normal, $-\sigma\Phi^{-1}(\mu)$ is the distance to a half space for which the error rate is $\mu$. In other words, the right hand side of the inequality is the same expression that appears in Equation~\ref{eqn:distscale}.
    
    So, among models with some fixed error rate $\errorvolume$, the most robust are the ones whose error set is a half space (as shown in Figure~\ref{fig:halfspace}). In Appendix~\ref{app:isoperimetric} we will give a more common statement of the Gaussian isoperimetric inequality along with a proof of the version presented here.
    
    \textbf{Comparing Neural Networks to the Isoperimetric Bound.} We evaluated these quantities for several models on the CIFAR-10 and ImageNet test sets.
    
    As in Section~\ref{sec:cleanimage}, we report an estimate of $\epsilon^*_q$. For each test image, we took 1,000 samples from the corresponding Gaussian and estimated $\epsilon^*_q$ using PGD with 200 steps on each sample and reported the median.
    
    We find that for the five models we considered, the relationship between our estimate of $\epsilon^*_q(E)$ and $\errorvolume$ is already close to optimal. This is visualized in Figure~\ref{fig:isoplot}. For CIFAR-10, adversarial training improves robustness to small perturbations, but the gains are primarily because error rates in Gaussian noise were improved. In particular, it is clear from the graph on the bottom left that adversarial training increases the $\sigma$ at which the error rate is 1\% on average. This shows that improved adversarial robustness results in improved robustness to large random perturbations, as the isoperimetric inequality says it \emph{must}.
     
    %Not all models or functions will be this close to optimal. As a simple example, if we took one of the CIFAR models shown in Figure~\ref{fig:isoplot} and modified it so that the model outputs an error whenever each coordinate of the input is an integer multiple of $10^{-6}$, the resulting model would have an error within $\sqrt{\frac12\cdot 10^{-6}\cdot\dim(\mathrm{CIFAR})}\approx 0.039$ of every point. In this case, adversarial examples \emph{would} be a distinct phenomenon from test performance, since $\epsilon^*_q(E)$ would be far from optimal.
    
    % Note that moving a point directly upward in Figure~\ref{fig:isoplot} amounts to rearranging the error set so that it has the same volume but a smaller surface area. This does not improve generalization, nor does it make it harder for an adversary to find an error. Furthermore, even if there is a way to rearrange errors to reduce the surface area, it is unclear what purpose this would serve unless the goal is \emph{only} to eliminate errors in some small $l_p$ ball.
 
\section{Evaluating Corruption Robustness}
\begin{figure*}[ht]
    \centering
    \includegraphics[width=0.8\linewidth]{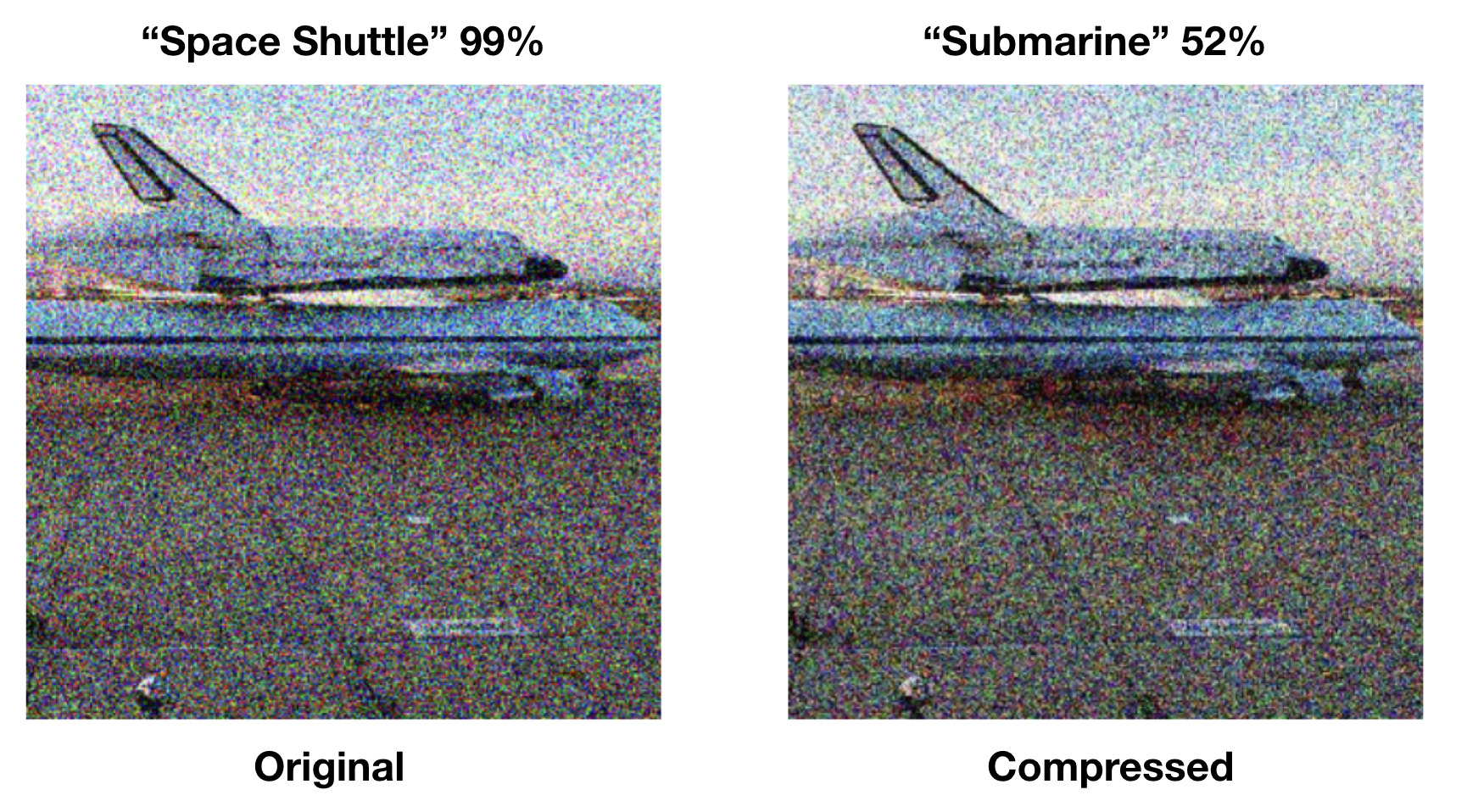}
    \caption{\label{fig:jpeg} Performance on the Imagenet-C corruptions may vary dramatically depending on whether or not the model is evaluated on the publicly released compressed images vs applying the corruptions directly in memory. For example, an InceptionV3 model trained with Gaussian data augmentation was 57\% accurate on the Gaussian-5 corruption when evaluated in memory (example image left). This same model was only 10\% accurate on the publicly released compressed images (example image right). The model prediction and confidence on each image is also shown. Note \textbf{the image on the right was not modified adversarially}, instead the drop in model performance is due entirely to subtle compression artifacts. This severe degradation in model performance is particularly surprising because differences between the compressed and uncompressed images are difficult to spot for a human. This demonstrates the extreme brittleness of neural networks to \emph{distributional shift}.}
\end{figure*}

    \label{sec:advtrain}
    The previous two sections show a relationship between adversarial robustness and one type of corruption robustness. This suggests that methods designed to improve adversarial robustness ought to also improve corruption robustness, and vice versa. In this section we investigate this relationship.

    \begin{figure*}[ht]
        \centering
        \includegraphics[width=.45\textwidth]{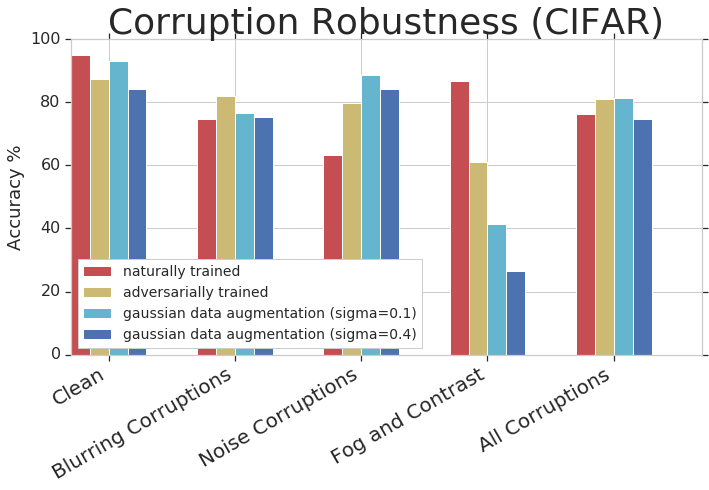}
        \hspace{.075\textwidth}
        \includegraphics[width=.45\textwidth]{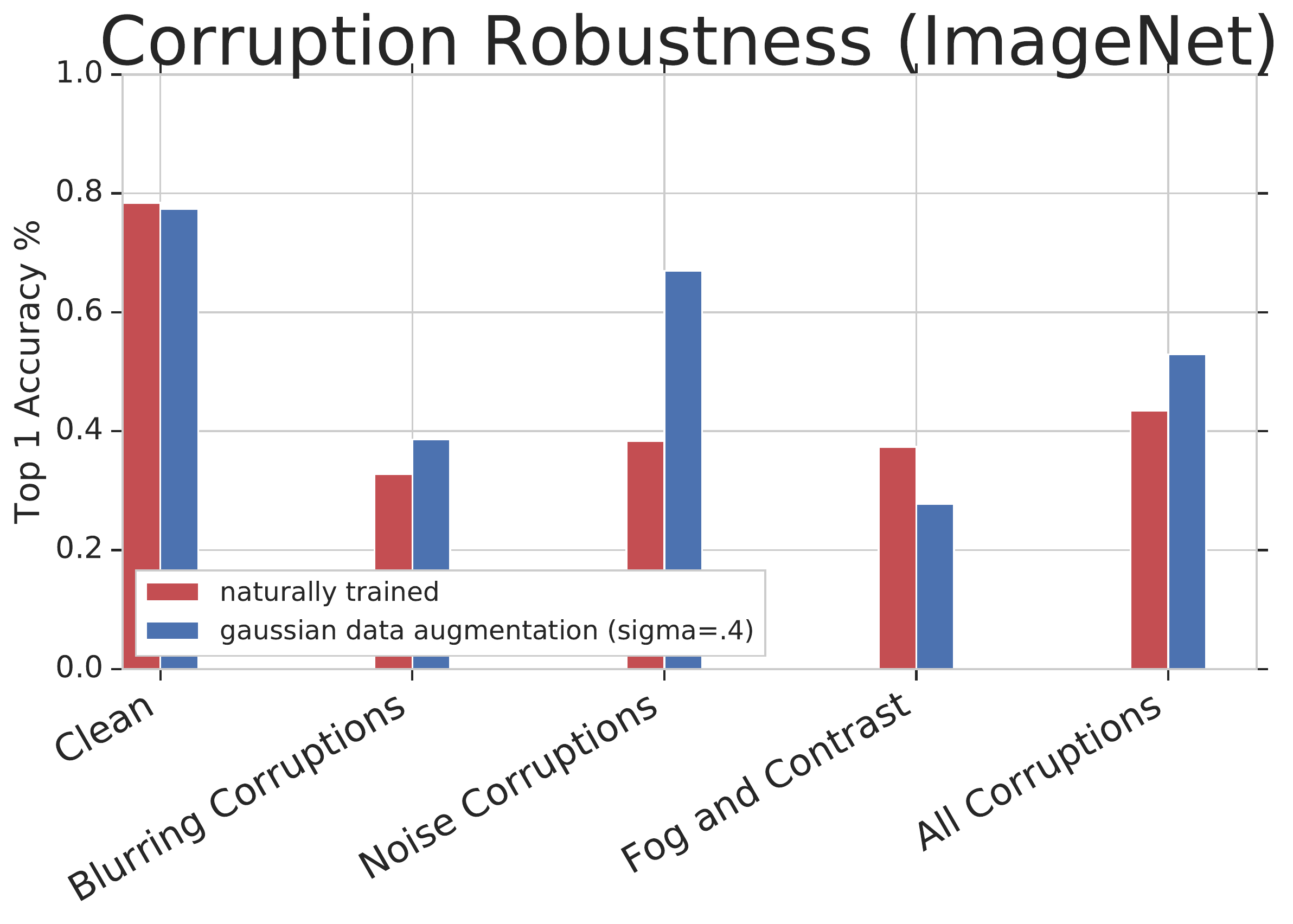}\\
        \vspace{1ex}
        \includegraphics[width=.45\textwidth]{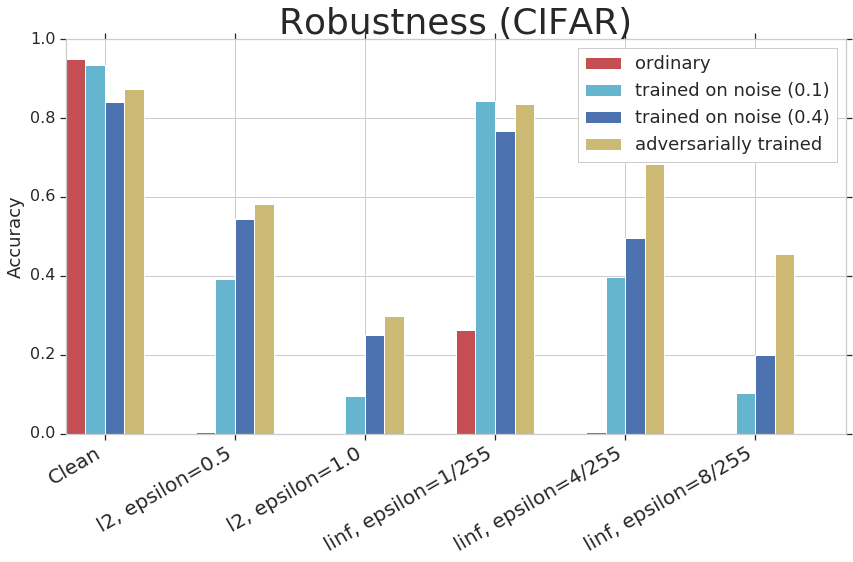}
        \hspace{.05\textwidth}
        \includegraphics[width=.45\textwidth]{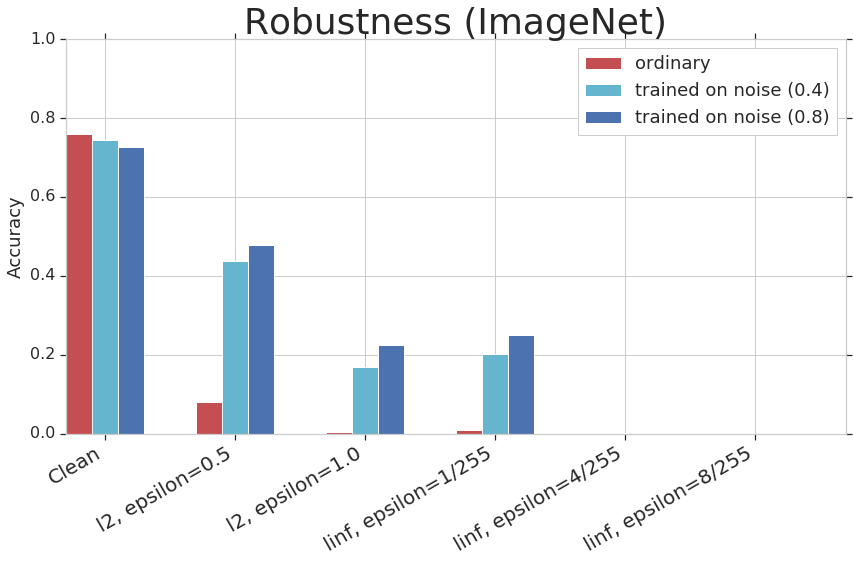}
        \caption{\label{fig:allnoises}The performance of the models we considered on the corruption robustness benchmark, together with our measurements of those models' robustness to small $l_p$ perturbations. For all the robustness tests we used PGD with 100 steps and a step size of $\epsilon/25$. The adversarially trained CIFAR-10 model is the open sourced model from \citet{madry2017advexamples}.}
    \end{figure*}
    
    We analyzed the performance of our models on the corruption robustness benchmark described in \citet{hendrycks2018benchmarking}. There are 15 different corruptions in this benchmark, each of which is tested at five different levels of severity. The results are summarized in Figure~\ref{fig:allnoises}, where we have aggregated the corruption types based on whether the ordinarily trained model did better or worse than the augmented models. We found a significant difference in performance on this benchmark when the model is evaluated on the compressed images provided with the benchmark rather than applying the corruptions in memory. (In this section we report performance on corruptions applied in-memory.) Figure~\ref{fig:jpeg} shows an example for the Gaussian-5 corruption, where performance degraded from 57\% accuracy (in memory) to 10\% accuracy (compressed images). Detailed results on both versions of this benchmark are presented in Appendix~\ref{sec:app_full_corr}.
    
    Gaussian data augmentation and adversarial training both improve the overall benchmark\footnote{In reporting overall performance on this benchmark, we omit the Gaussian noise corruption.}, which requires averaging the performance across all corruptions, and the results were quite close. Adversarial training helped more with blurring corruptions and Gaussian data augmentation helped more with noise corruptions. Interestingly, both methods performed much worse than the clean model on the fog and contrast corruptions. For example, the adversarially trained model was 55\% accurate on the most severe contrast corruption compared to 85\% for the clean model. Note that \citet{hendrycks2018benchmarking} also observed that adversarial training improves robustness on this benchmark on Tiny ImageNet.
    
    The fact that adversarial training is so successful against the noise corruptions further supports the connection we have been describing. For other corruptions, the relationship is more complicated, and it would be interesting to explore this in future work.
    %\todo{take a lookat efficient defenses adversarial attack aisec'17}
    
    We also evaluated these two augmentation methods on standard measures of $l_p$ robustness. We see a similar story there: while adversarial training performs better, Gaussian data augmentation does improve adversarial robustness as well. Gaussian data augmenation has been proposed as an adversarial defense in prior work \cite{zantedeschi2017efficient}. Here we evaluate this method not to propose it as a novel defense but to provide further evidence of the connection between adversarial and corruption robustness.

    We also considered the MNIST adversarially trained model from \citet{madry2017advexamples}, and found it to be a special case where robustness to small perturbations was increased while generalization in noise was not improved (see Appendix~\ref{app:mnist}). This is because this model violates the linearity assumption discussed in Section~\ref{sec:cleanimage}.
    
\begin{figure*}[ht]
    \centering
    \includegraphics[width=0.4\linewidth]{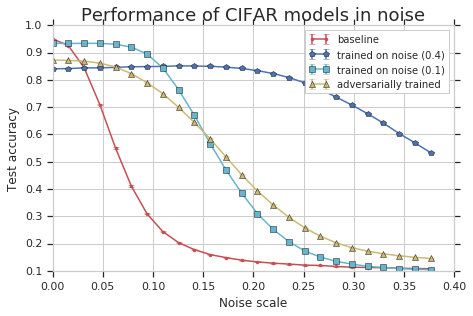}
    \includegraphics[width=0.5\linewidth]{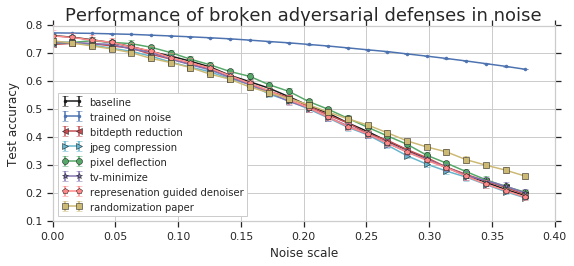}
    \caption{\label{fig:brokendefenses}\textit{(Left)} The performance in Gaussian noise of the CIFAR models described in this paper. \textit{(Right)} The performance in Gaussian noise of several previously published defenses for ImageNet, along with an Imagenet model trained on Gaussian noise at $\sigma=0.4$ for comparison. For each point we ran ten trials; the error bars show one standard deviation. All of these defenses are now known not to improve adversarial robustness \citep{athalye2018obfuscated}. The defense strategies include bitdepth reduction \citep{guo2017countering}, JPEG compression \citep{guo2017countering, dziugaite2016study, liu2018feature, aydemir2018effects, das2018shield,das2017keeping}, Pixel Deflection \citep{prakash2018deflecting}, total variance minimization \citep{guo2017countering}, respresentation-guided denoising \citep{liao2018defense}, and random resizing and random padding of the input image \citep{xie2017mitigating}.}
    \end{figure*}

    \textbf{Corruption Robustness as a Sanity Check for Defenses.} We also analyzed the performance several previously published adversarial defense strategies in Gaussian noise. These methods have already been shown to result in vanishing gradients, which causes standard optimization procedures to fail to find errors, rather than actually improving adversarial robustness \citep{athalye2018obfuscated}. We find that these methods also show no improvement in Gaussian noise. The results are shown in Figure~\ref{fig:brokendefenses}. Had these prior defenses performed an analysis like this, they would have been able to determine that their methods relied on vanishing gradients and fail to improve robustness.%Given how easy it is for a method to show improved robustness to standard optimization procedures without changing the decision boundary in any meaningful way, we strongly recommend that future defense efforts evaluate corruption robustness. The current standard practice of evaluating solely on gradient-based attack algorithms is making progress more difficult to measure.
    
    \textbf{Obtaining Zero Test Error in Noise is Nontrivial.} It is important to note that applying Gaussian data augmentation does not reduce error rates in Gaussian noise to zero. For example, we performed Gaussian data augmentation on CIFAR-10 at $\sigma=.15$ and obtained 99.9\% training accuracy but 77.5\% test accuracy in the same noise distribution. (For comparison, the naturally trained obtains 95\% clean test accuracy.) Previous work \citep{dodge2017study} has also observed that obtaining perfect generalization in large Gaussian noise is nontrivial. This mirrors \citet{schmidt2018adversarially}, which found that adversarial robustness did not generalize to the test set, providing yet another similarity between adversarial and corruption robustness. This is perhaps not surprising given that error rates on the \emph{clean} test set are also non-zero. Although the model is in some sense ``superhuman'' with respect to clean test accuracy, it still makes many mistakes on the clean test set that a human would never make. We collected some examples in Appendix~\ref{app:nonhuman}. More detailed results on training and testing in noise can be found in Appendices \ref{app:noiseresults} and \ref{app:noisehist}.
    
\section{Conclusion}
    This paper investigates whether we should be surprised to find adversarial examples as close as we do, given the error rates we observe in corrupted image distributions. After running several experiments, we argue that the answer to this question is no. Specifically:
    
    \begin{enumerate}
        \item The nearby errors we can find show up at the same distance scales we would expect from a linear model with the same corruption robustness.
        
        \item Concentration of measure shows that a non-zero error rate in Gaussian noise \emph{logically implies} the existence of small adversarial perturbations of noisy images. 
        %Using Gaussian perturbations of a single image to measure both adversarial robustness and test performance, we find a bound that even the best-behaved decision boundary must satisfy, and we find that today's models are already close to meeting this bound.
        
        \item Finally, training procedures designed to improve adversarial robustness also improve many types of corruption robustness, and training on Gaussian noise moderately improves adversarial robustness.
    \end{enumerate}
    
    %These results all point to the conclusion that the existence of adversarial examples is due simply to fact that, in high dimensions, even a relatively small error set can be close to most samples from the image distribution; there is no need to invoke any strange properties of the shape of the error set itself. This is important for adversarial defense design. If adversarial examples arose from a badly behaved decision boundary, then it would make sense to design defenses which attempt to smooth out the decision boundary in some way. Currently there is a considerable subset of the adversarial defense literature which develops methods that would remove any small ``pockets'' of errors but which don't improve model generalization. One example is \citet{xie2017mitigating} which proposes randomly resizing the input to the network as a defense strategy. Unfortunately, this defense, like many others, has been shown to be ineffective against stronger adversaries \citep{carlini2017adversarial, carlini2017towards, athalye2018obfuscated}.
    
    In light of this, we believe it would be beneficial for the adversarial defense literature to start reporting generalization to distributional shift, such as the common corruption benchmark introduced in \citet{hendrycks2018benchmarking}, in addition to empirical estimates of adversarial robustness. There are several reasons for this recommendation.%\todo{It would be good to warn about accidental cheating}
    
    First, a varied suite of corruptions can expose failure modes of a model that we might otherwise miss. For example, we found that adversarial training significantly degraded performance on the fog and contrast corruptions despite improving small perturbation robustness. In particular, performance on constrast-5 dropped to 55.3\% accuracy vs 85.7\% for the vanilla model (see Appendix~\ref{sec:app_full_corr} for more details). 
    
    Second, measuring corruption robustness is significantly easier than measuring adversarial robustness --- computing adversarial robustness perfectly requires solving an NP-hard problem for every point in the test set \citep{katz2017reluplex}. Since \citet{Szegedy14}, hundreds of adversarial defense papers have been published. To our knowledge, only one \citep{madry2017advexamples} has reported robustness numbers which were confirmed by a third party. We believe the difficulty of measuring robustness under the usual definition has contributed to this unproductive situation.
    
    Third, all of the failed defense strategies we examined also failed to improve performance in Gaussian noise. For this reason, we should be highly skeptical of defense strategies that only claim improved $l_p$ robustness but are unable to demonstrate robustness to distributional shift.
    
    Finally, if the goal is improving the security of our models in adversarial settings, errors on corrupted images already imply that our models are not secure. Until our models are perfectly robust in the presence of average-case corruptions, they will not be robust in worst-case settings. %The usefulness of $l_p$ robustness in realistic threat models is limited when attackers are not constrained to making small modifications.\todo{While I agree with this, I don't think it's something that fits the message of the paper.}
    
    The communities of researchers studying adversarial and corruption robustness seem to be attacking essentially the same problem in two different ways. We believe that the corruption robustness problem is also interesting independently of its connection to adversarial examples, and we hope that the results presented here will encourage more collaboration between these two communities. % to attack it more seriously.\todo{"the community" which? "it" what?}

\bibliographystyle{iclr2019_conference}
\bibliography{test_error}

\newpage
\onecolumn
\appendix

\section{Training Details}

    \label{app:trainingdetails}
    \textbf{Models trained on CIFAR-10.} We trained the Wide-ResNet-28-10 model~\citep{zagoruyko2016wide} using standard data augmentation of flips, horizontal shifts and crops in addition to Gaussian noise independently sampled for each image in every minibatch. The models were trained with the open-source code by \citet{cubuk2018autoaugment} for 200 epochs, using the same hyperparameters which we summarize here: a weight decay of 5e-4, learning rate of 0.1, batch size of 128. The learning rate was decayed by a factor of 0.2 at epochs 60, 120, 160.
        
    \textbf{Models trained on ImageNet.} The Inception v3 model~\citep{szegedy16} was trained with a learning rate of 1.6, batch size of 4096, and weight decay of 8e-5. During training, Gaussian noise was independently sampled for each image in every minibatch. The models were trained  for 130 epochs, where the learning rate was decayed by a factor of 0.975 every epoch. Learning rate was linearly increased from 0 to the value of 1.6 over the first 10 epochs.
        
\section{Full Corruption Robustness Results}
\label{sec:app_full_corr}
In this section we examine the corruption robustness of both adversarially trained models and models trained with Gaussian data augmentation. Full results are shown in Tables~\ref{tab:cor_robust},~\ref{tab:cor_robust_cifar}. We highlight several interesting findings from these experiments. %Accuracies on all corruptions and severities can be found in the csv files included in the supplementary material. 

\begin{figure*}[ht]
    \centering
    \includegraphics[width=0.7\linewidth]{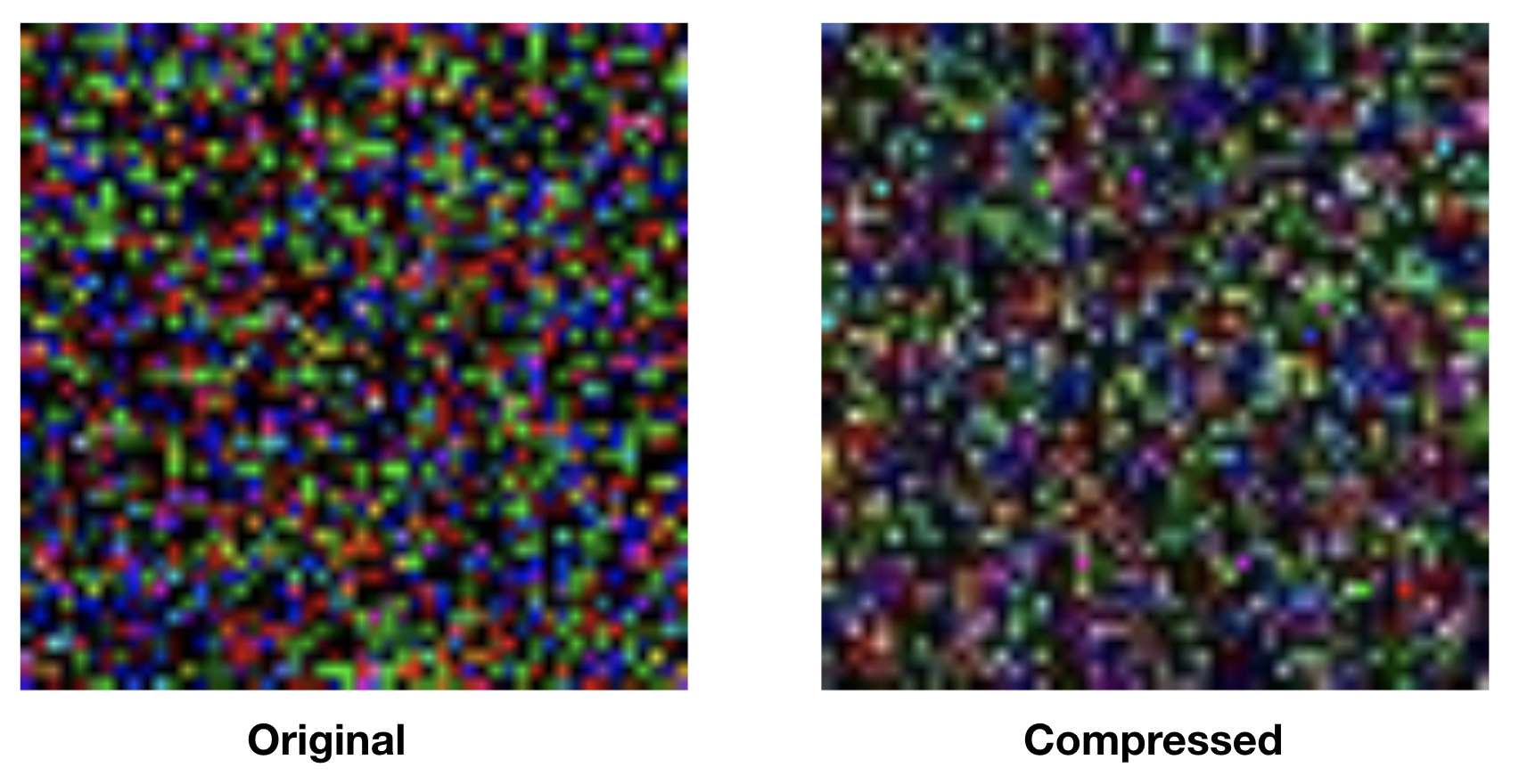}
    \caption{\label{fig:jpeg_noise} Visualizing the effects of jpeg compression on white noise. The subtle difference between the compressed and uncompressed images is enough to degrade model performance on several of the Imagenet-C corruptions.}
\end{figure*}

\begin{itemize}
\item On CIFAR-10-C, Gaussian data augmentation outperforms adversarial training on the overall benchmark. However, adversarial training is better on all of the blurring corruptions.
\item The publicly released Imagenet-C dataset as .jpeg files is significantly harder than the same dataset when the corruptions are applied in memory. It appears that this is due to additional artifacts added to the image from the JPEG compression algorithm (see Figure~\ref{fig:jpeg_noise}). Future work should make care of this distinction when comparing the performance of their methods, in particular we note that the results in \citep{geirhos2018imagenet, hendrycks2018benchmarking} were both evaluated on the jpeg files. 
\item Both adversarial training and Gaussian data augmenation significantly degrade performance on the severe fog and constrast corruptions (Tables~\ref{tab:fog_imnet},~\ref{tab:fog_cifar}). This highlights the importance of evaluating on a broad suite of corruptions as simply evaluating on worst-case $l_p$ perturbations or random noise will not expose all failings of a model. This also highlights the need for developing methods that improve robustness to all corruptions. Towards this end the exciting new ``Stylized ImageNet''\citep{geirhos2018imagenet} data augmentation process achieves moderate improvements on all corruptions, at least on the publicly released .jpeg files. 
\end{itemize}

\begin{table}[]
    \caption{\label{tab:cor_robust} Measuring the improvements of Gaussian data augmentation on corruption robustness for Imagenet-C. For this table we evaluate both on corruptions in memory to the existing Imagenet validation set using the code at \url{https://github.com/hendrycks/robustness} and on the compressed version of the dataset  from \url{https://drive.google.com/drive/folders/1HDVw6CmX3HiG0ODFtI75iIfBDxSiSz2K?usp=sharing}. We found that model performance when the corruption was applied in memory is higher than performance on the publicly released .jpeg files that already have the corruptions applied to them. Unfortunately, we were unable to evaluate all corruptions due to issues installing some of the dependencies, these are marked with a ?. All numbers are model accuracies averaged over the 5 corruption severities. }
\centering

\begin{tabular}{lllllllll}
  & & \multicolumn{3}{c}{Noise} & \multicolumn{4}{c}{Blur} \\
\cline{3-5} \cline{6-9}
Training & All & Gaussian & Shot    & Impulse  & Defocus  & Glass     & Motion & Zoom    \\
\hline
Vanilla InceptionV3 & 45.0 & 40.3 & 38.7  & 38.0 & 40.3  & 26.4 & ?     & 31.6 \\
Gaussian ($\sigma=0.4$) & \textbf{52.6} & \textbf{67.5} & \textbf{67.5} & \textbf{66.4} & \textbf{43.4} & \textbf{39.4}   & ?     & \textbf{33.0} \\

  & \multicolumn{4}{c}{Weather} & \multicolumn{4}{c}{Digital} \\
  \cline{2-5} \cline{6-9}
Training & Snow & Frost & Fog & Brightness & Contrast & Elastic & Pixelate & JPEG \\
\hline
Vanilla InceptionV3 & ? & ? & \textbf{60.0} & 68.6   & \textbf{45.2} & 46.8 & 42.8 & 56.2 \\
Gaussian ($\sigma=0.4$) & ? & ? & 54.0 & \textbf{68.8}  & 39.0  & \textbf{51.6} & \textbf{51.8} & \textbf{63.6}  \\

  & & \multicolumn{3}{c}{Noise (Compressed)} & \multicolumn{4}{c}{Blur (Compressed)} \\
\cline{3-5} \cline{6-9}
Training & All & Gaussian & Shot    & Impulse  & Defocus  & Glass     & Motion & Zoom    \\
\hline
 Vanilla InceptionV3 & 38.8 & 36.6       & 34.3 & 34.7 & 31.1  & 19.3    & 35.3 & 30.1  \\
 Gaussian ($\sigma=0.4$) & \textbf{42.7} & \textbf{40.3}  & \textbf{38.8} & \textbf{37.7} & \textbf{32.9} & \textbf{29.8} & \textbf{35.3} & \textbf{33.1} \\
 
   & \multicolumn{4}{c}{Weather (Compressed)} & \multicolumn{4}{c}{Digital (Compressed)} \\
  \cline{2-5} \cline{6-9}
Training & Snow & Frost & Fog & Brightness & Contrast & Elastic & Pixelate & JPEG \\
\hline
Vanilla InceptionV3 & 33.1 & 34.0  & \textbf{52.4}  & 66.0 & \textbf{35.9} & 47.8 & 38.2 & 50.9  \\
Gaussian ($\sigma=0.4$) & \textbf{36.6} & \textbf{43.5} & 52.3 & \textbf{67.1}  & 35.8  & \textbf{52.2} & \textbf{47.0} & \textbf{55.5}  \\

\end{tabular}
\end{table}

%\begin{table}[]
%    \caption{\label{tab:cor_robust_tf} Measuring the improvements of Gaussian data augmentation on corruption robustness for Imagenet-C. For this table we evaluate on the publicly released .jpeg files found at \url{https://drive.google.com/drive/folders/1HDVw6CmX3HiG0ODFtI75iIfBDxSiSz2K?usp=sharing}. We found that model performance on these .jpeg files was lower than when the corruption is applied in memory. This is likely due to the JPEG compression algorithm further corrupting the image. All numbers are model accuracies averaged over the 5 corruption severities.}
%    \centering
%\begin{tabular}{lllllllll}
%  & & \multicolumn{3}{c}{Noise} & \multicolumn{4}{c}{Blur} \\
%\cline{3-5} \cline{6-9}
%Training & All & Gaussian & Shot    & Impulse  & Defocus  & Glass     & Motion & Zoom    \\
%\hline
% Vanilla InceptionV3 & 38.8 & 36.6       & 34.3 & 34.7 & 31.1  & 19.3    & 35.3 & 30.1  \\
% Gaussian ($\sigma=0.4$) & \textbf{42.7} & \textbf{40.3}  & \textbf{38.8} & \textbf{37.7} & \textbf{32.9} & \textbf{29.8} & \textbf{35.3} & \textbf{33.1} \\
 
%   & \multicolumn{4}{c}{Weather} & \multicolumn{4}{c}{Digital} \\
%  \cline{2-5} \cline{6-9}
%Training & Snow & Frost & Fog & Brightness & Contrast & Elastic & Pixelate & JPEG \\
%\hline
%Vanilla InceptionV3 & 33.1 & 34.0  & \textbf{52.4}  & 66.0 & \textbf{35.9} & 47.8 & 38.2 & 50.9  \\
%Gaussian ($\sigma=0.4$) & \textbf{36.6} & \textbf{43.5} & 52.3 & \textbf{67.1}  & 35.8  & \textbf{52.2} & \textbf{47.0} & \textbf{55.5}  \\

%\end{tabular}
%\end{table}

\begin{table}[]
    \caption{\label{tab:cor_robust_cifar} Comparing the corruption robustness of adversarial training and Gaussian data augmentation on the CIFAR-10-C dataset. For this table we evaluate on the publicly release .npy files found at \url{https://drive.google.com/drive/folders/1HDVw6CmX3HiG0ODFtI75iIfBDxSiSz2K?usp=sharing}. Unlike the Imagenet-C dataset which was released as .jpeg files, there was no additional noise applied when saving the images as .npy files. All numbers are model accuracies averaged over the 5 corruption severities.}
    \centering
\begin{tabular}{lllllllll}
%\multicolumn{9}{c}{\textbf{CIFAR-10 Corruption Robustness}} \\
  & & \multicolumn{3}{c}{Noise} & \multicolumn{4}{c}{Digital} \\
\cline{3-5} \cline{6-9}

Training & All & Speckle & Shot    & Impulse & Contrast & Elastic & Pixelate & JPEG    \\
\hline
Vanilla Wide-ResNet-28-10 & 76.3 & 62.8  & 59.3 & 53.3 & \textbf{92.2}  & \textbf{84.8}  & 74.0  & 77.2 \\
Adversarialy Trained & 80.9 & 81.8 & 82.8 & 68.8 & 77.0  & 81.8 & 85.3   & 85.4 \\
Gaussian ($\sigma=0.1$) & \textbf{81.2} & \textbf{91.1} & \textbf{91.8} & 81.5 & 58.9  & 82.2 & \textbf{89.0}  & \textbf{90.0} \\
Gaussian ($\sigma=0.4$) & 74.7 & 84.6  & 84.6 & \textbf{84.5} & 41.5  & 75.4 & 81.2  & 82.9 \\

  & & \multicolumn{3}{c}{Weather} & \multicolumn{4}{c}{Blur} \\
\cline{3-5} \cline{6-9}

Training & Snow    & Fog     & Brightness & Defocus & Glass   & Motion  & Zoom    & Gaussian \\
\hline
Vanilla Wide-ResNet-28-10 & 83.3 & \textbf{90.4}  & \textbf{94.0}    & \textbf{85.5} & 51.1 & \textbf{81.2} & 79.9 & 75.3  \\
Adversarialy Trained & 82.6  & 72.7 & 87.1    & 83.5 & \textbf{80.2}  & 80.5 & \textbf{82.8}  & \textbf{82.1}  \\
Gaussian ($\sigma=0.1$) & \textbf{87.3}  & 71.5 & 91.8    & 80.0  & 79.6 & 71.6  & 77.2  & 74.2   \\
Gaussian ($\sigma=0.4$) & 78.0 & 51.8 & 80.1     & 77.0 & 77.9 & 72.0 & 74.8  & 74.4 

\end{tabular}
\end{table}

\begin{table}[h]

\caption{\label{tab:fog_imnet}Detailed results for the fog and contrast corruptions on ImageNet-C highlighting the effect of the severity on both the compressed and uncompressed versions of the data. When the corruption is applied in memory, Gaussian data augmentation degrades performance in comparison to a clean model. However, when evaluating on the compressed version of this dataset this degradation in comparison to the clean model is minimized.}
\centering
\begin{tabular}{l|rr|rr}
\toprule
 corruption &   clean & trained on noise & clean (compressed) & trained on noise (compressed) \\
\midrule
 contrast-1 &  \textbf{68.198} &  66.528 &   62.502 &  \textbf{63.876} \\
 contrast-2 &  \textbf{63.392} &  60.634 &   55.626 &  \textbf{57.308} \\
 contrast-3 &  \textbf{53.878} &   47.57 &   42.024 &  \textbf{42.434} \\
 contrast-4 &  \textbf{30.698} &   17.34 &   \textbf{16.172} &  13.122 \\
 contrast-5 &   \textbf{9.746} &   2.798 &    \textbf{3.362} &    2.07 \\
      fog-1 &  \textbf{67.274} &  65.148 &   61.334 &   \textbf{62.91} \\
      fog-2 &   \textbf{63.77} &  60.398 &    56.51 &  \textbf{57.746} \\
      fog-3 &   \textbf{59.51} &  53.752 &   51.188 &  \textbf{51.292} \\
      fog-4 &  \textbf{58.098} &   51.34 &   \textbf{50.064} &  49.324 \\
      fog-5 &  \textbf{50.996} &  39.586 &   \textbf{42.874} &   40.34 \\
\bottomrule
\end{tabular}
\end{table}

\begin{table}[h]
\caption{\label{tab:fog_cifar}Detailed results for the fog and contrast corruptions on CIFAR-10-C. Both adversarial training and Gaussian data augmenation significantly degrade performance on these corruptions.}
\centering
\begin{tabular}{l|rrrr}
\toprule
 corruption &  clean &    adv &  Gaussian (0.1) & Gaussian (0.4) \\
\midrule
 contrast-0 &  \textbf{94.73} &  86.65 &  90.51 &  76.45 \\
 contrast-1 &  \textbf{94.22} &  84.59 &  77.12 &  50.71 \\
 contrast-2 &  \textbf{93.67} &  82.09 &  63.71 &  36.49 \\
 contrast-3 &  \textbf{92.51} &  76.40 &  43.74 &  25.96 \\
 contrast-4 &  \textbf{85.66} &  55.29 &  19.36 &  17.98 \\
      fog-0 &  \textbf{94.90} &  86.75 &  91.53 &  78.87 \\
      fog-1 &  \textbf{94.75} &  84.65 &  86.05 &  65.50 \\
      fog-2 &  \textbf{93.98} &  79.16 &  77.93 &  51.99 \\
      fog-3 &  \textbf{91.69} &  68.41 &  64.11 &  38.62 \\
      fog-4 &  \textbf{76.58} &  44.17 &  38.01 &  24.04 \\
\bottomrule
\end{tabular}
\end{table}

\section{Training and Testing on Gaussian Noise}
    \label{app:noiseresults}
    In Section~6, we mentioned that it is not trivial to learn the distribution of noisy images simply by augmenting the training data distribution. In Tables \ref{tab:cifarnoise} and \ref{tab:imagenetnoise} we present more information about the performance of the models we trained and tested on various scales of Gaussian noise.
    
    \begin{table*}
    \caption{\label{tab:cifarnoise}Wide ResNet-28-10~\citep{zagoruyko2016wide} trained and tested on CIFAR-10 with Gaussian noise with standard deviation $\sigma$.}
    \centering
    \begin{tabular}{lrrrrrr}
      \hline
        $\sigma$ & 0.00625 & 0.0125 & 0.025 &  0.075 &  0.15 & 0.25 \\
        \hline
        Training Accuracy & 100\%  & 100\%  & 100\%  & 100\%  & 99.9\% & 99.4\% \\
        Test Accuracy     & 96.0\% & 95.5\% & 94.8\% & 90.4\% & 77.5\% & 62.2\% \\
      \hline
    \end{tabular}
    \end{table*}
    
    \begin{table*}
    \caption{\label{tab:imagenetnoise}The models from Section~6 trained and tested on ImageNet with Gaussian noise with standard deviation $\sigma$; the column labeled 0 refers to a model trained only on clean images.}
    \centering
    \begin{tabular}{lrrrrrrrr}
      \hline
        $\sigma$ & 0 & 0.1 & 0.2 & 0.4 & 0.6 & 0.8 \\
        \hline
        Clean Training Accuracy & 91.5\% & 90.8\% & 89.9\% & 87.7\% & 86.1\% & 84.6\% \\
        Clean Test Accuracy     & 75.9\% & 75.5\% & 75.2\% & 74.2\% & 73.3\% & 72.4\% \\
        \hline
        Noisy Training Accuracy & $-$    & 89.0\% & 85.7\% & 78.3\% & 71.7\% & 65.2\% \\
        Noisy Test Accuracy     & $-$    & 73.9\% & 70.9\% & 65.2\% & 59.7\% & 54.0\% \\
      \hline
    \end{tabular}
    \end{table*}

\section{Results on MNIST}
    \label{app:mnist}
    MNIST is a special case when it comes to the relationship between small adversarial perturbations and generalization in noise. Indeed prior has already observed that an MNIST model can trivially become robust to small $l_\infty$ perturbations by learning to threshold the input \citep{schmidt2018adversarially}, and observed that the model from \citet{madry2017advexamples} indeed seems to do this. When we investigated this model in different noise distributions we found it generalizes worse than a naturally trained model, results are shown in Table~\ref{tab:mnistnoises}. Given that it is possible for a defense to overfit to a particular $l_p$ metric, future work would be strengthened by demonstrating improved generalization outside the natural data distribution. 

    \begin{table*}
    \caption{\label{tab:mnistnoises}The performance of ordinarily and adversarially trained MNIST models on various noise distributions.}
    \centering
    \begin{tabular}{rccccc}
      \hline
               & & Pepper & Gaussian & stAdv & PCA-100 \\
               & Clean & $p=0.2$ & $\sigma=0.3$ & $\sigma=1.0$ & $\sigma=0.3$ \\
      Model    & Accuracy & Accuracy   & Accuracy     & Accuracy   & Accuracy\\
      \hline
        Clean & 99.2\% & 81.4\% & 96.9\% & 89.5\% & 63.3\% \\
        Adv   & 98.4\% & 27.5\% & 78.2\% & 93.2\% & 47.1\% \\
      \hline
    \end{tabular}
    \end{table*}

    Here we provide more detail for the noise distributions we used to evaluate the MNIST model. The stAdv attack defines a flow field over the pixels of the image and shifts the pixels according to this flow. The field is parameterized by a latent $Z$. When we measure accuracy against our randomized variant of this attack, we randomly sample $Z$ from a multivariate Gaussian distribution with standard deviation $\sigma$. To implement this attack we used the open sourced code from \citet{xiao2018spatially}. PCA-100 noise first samples noise from a Gaussian distribution $\mathcal{N}(0, \sigma)$, and then projects this noise onto the first 100 PCA components of the data.% For ImageNet, the input dimension is too large to perform a PCA decomposition on the entire dataset. So we first perform a PCA decomposition on 30x30x1 patches taken from different color channels of the data. To general the noise we first sample from a 900 dimensional Gaussian, then project this into the basis spanned by the top 100 PCA components, then finally tile this projects to the full 299x299 dimension of the input. Each color channel is constructed independently in this fashion.  %Note that this shows that gradient descent, or white box access is not required for a successful attack, but modifying the image at random is sufficient to cause an error. In the future we recommend adversarial attack papers to consider randomized variants of their attack algorithms, as only considering small worst-case perturbations can lead to a misconception that model errors can only be found via an optimization procedure.

\section{The Gaussian Isoperimetric Inequality}
    \label{app:isoperimetric}
    Here we will discuss the Gaussian isoperimetric inequality more thoroughly than we did in the text. We will present some of the geometric intuition behind the theorem, and in the end we will show how the version quoted in the text follows from the form in which the inequality is usually stated.
    
    The historically earliest version of the isoperimetric inequality, and probably the easiest to understand, is about areas of subsets of the plane and has nothing to do with Gaussians at all. It is concerned with the following problem: among all measurable subsets of the plane with area $A$, which ones have the smallest possible perimeter?\footnote{The name ``isoperimetric'' comes from a different, but completely equivalent, way of stating the question: among all sets with the same fixed perimeter, which ones have the largest possible area?} One picture to keep in mind is to imagine that you are required to fence off some region of the plane with area $A$ and you would like to use as little fence as possible. The isoperimetric inequality says that the sets which are most ``efficient'' in this sense are balls.
    
    Some care needs to be taken with the definition of the word ``perimeter'' here --- what do we mean by the perimeter of some arbitrary subset of $\mathbb{R}^2$? The definition that we will use involves the concept of the $\epsilon$-boundary measure we discussed in the text. For any set $E$ and any $\epsilon>0$, recall that we defined the \emph{$\epsilon$-extension} of $E$, written $E_\epsilon$, to be the set of all points which are within $\epsilon$ of a point in $E$; writing $A(E)$ for the area of $E$, we then define the perimeter of $E$ to be \[\mathrm{surf}(E):=\liminf_{\epsilon\to 0}\frac1\epsilon\left(A(E_\epsilon)-A(E)\right).\] A good way to convince yourself that this is reasonable is to notice that, for small $\epsilon$, $E_\epsilon-E$ looks like a small band around the perimeter of $E$ with width $\epsilon$. The isoperimetric inequality can then be formally expressed as giving a bound on the quantity inside the limit in terms of what it would be for a ball. (This is slightly stronger than just bounding the perimeter, that is, bounding the limit itself, but this stronger version is still true.) That is, for any measurable set $E\subseteq\mathbb{R}^2$, \[\frac1\epsilon(A(E_\epsilon)-A(E))\ge 2\sqrt{\pi A(E)}+\epsilon\pi.\] It is a good exercise to check that we have equality here when $E$ is a ball.
    
    There are many generalizations of the isoperimetric inequality. For example, balls are also the subsets in $\mathbb{R}^n$ which have minimal surface area for a given fixed volume, and the corresponding set on the surface of a sphere is a ``spherical cap,'' the set of points inside a circle drawn on the surface of the sphere. The version we are most concerned with in this paper is the generalization to a Gaussian distribution. Rather than trying to relate the volume of $E$ to the volume of $E_\epsilon$, the Gaussian isoperimetric inequality is about the relationship between the \emph{probability} that a random sample from the Gaussian distribution lands in $E$ or $E_\epsilon$. Other than this, though, the question we are trying to answer is the same: for a given probability $p$, among all sets $E$ for which the probability of landing in $E$ is $p$, when is the probability of landing in $E_\epsilon$ as small as possible?
    
    \begin{figure}[t]
        \centering
        \begin{subfigure}{0.4\columnwidth}
            \includegraphics[width=\textwidth]{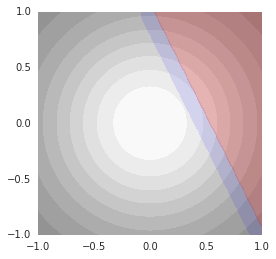}
        \end{subfigure}
        \begin{subfigure}{0.4\columnwidth}
            \includegraphics[width=\textwidth]{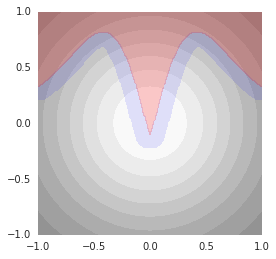}
        \end{subfigure}
        \caption{The Gaussian isoperimetric inequality relates the amount of probability mass contained in a set $E$ to the amount contained in its $\epsilon$-extension $E_\epsilon$. A sample from the Gaussian is equally likely to land in the pink set on the left or the pink set on the right, but the set on the right has a larger $\epsilon$-extension. The Gaussian isoperimetric inequality says that the sets with the smallest possible $\epsilon$-extensions are half spaces.}
        \label{fig:2d_iso}
    \end{figure}
    
    The Gaussian isoperimetric inequality says that the sets that do this are half spaces. (See Figure~\ref{fig:2d_iso}.) Just as we did in the plane, it is convenient to express this as a bound on the probability of landing in $E_\epsilon$ for an arbitrary measurable set $E$. This can be stated as follows:
    
    \begin{thm}
        Consider the standard normal distribution $q$ on $\mathbb{R}^n$, and let $E$ be a measurable subset of $\mathbb{R}^n$. Write \[\Phi(t)=\frac{1}{\sqrt{2\pi}} \int_{-\infty}^t \exp(x^2/2) dx,\] the cdf of the one-variable standard normal distribution.
    
        For a measurable subset $E\subseteq\mathbb{R}^n$, write $\alpha(E)=\Phi^{-1}(\errorvolume)$. Then for any $\epsilon\ge 0$, \[\errorboundary\ge\Phi(\alpha(E)+\epsilon).\]
    \end{thm}
    
    The version we stated in the text involved $\epsilon^*_q(E)$, the median distance from a random sample from $q$ to the closest point in $E$. This is the same as the smallest $\epsilon$ for which $\errorboundary=\frac12$. So, when $\epsilon=\epsilon^*_q(E)$, the left-hand side of the Gaussian isoperimetric inequality is $\frac12$, giving us that $\Phi(\alpha+\epsilon^*_q(E))\le\frac12$.
    
    Since $\Phi^{-1}$ is a strictly increasing function, applying it to both sides preserves the direction of this inequality. But $\Phi^{-1}(\frac12)=0$, so we in fact have that $\epsilon^*_q(E)\le -\alpha$, which is the statement we wanted.
    
% \section{Distances to the Decision Boundary}
%     \label{app:pgdhist}
%     Using naturally trained models, we computed the distance to the nearest error we could find using PGD large number of test images from both CIFAR-10 and ImageNet. A histogram is shown in Figure~\ref{fig:pgdhist}. Because the test error is much higher for the ImageNet model, its histogram contains more zeroes. Despite this, though, the average distance to the decision boundary is much larger for ImageNet (0.203) than it is for CIFAR (0.115).
%     \begin{figure}
%         \centering
%         \includegraphics[width=0.7\textwidth]{pgd_hist.png}
%         \caption{\label{fig:pgdhist}A histogram of distances to the nearest error found using PGD for CIFAR-10 and ImageNet.}
%     \end{figure}

\section{Visualizing the Optimal Curves}
\label{app:curve_vis}
\begin{figure*}
        \centering
        \begin{subfigure}{0.7\textwidth}
            \includegraphics[width=\textwidth]{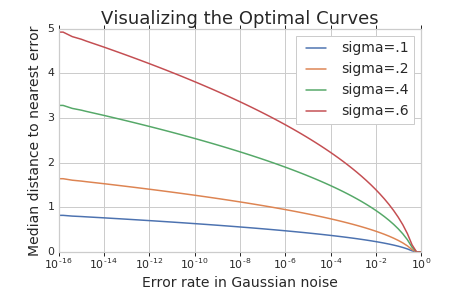}
        \end{subfigure}
        \begin{subfigure}{1.0\textwidth}
            \includegraphics[width=\textwidth]{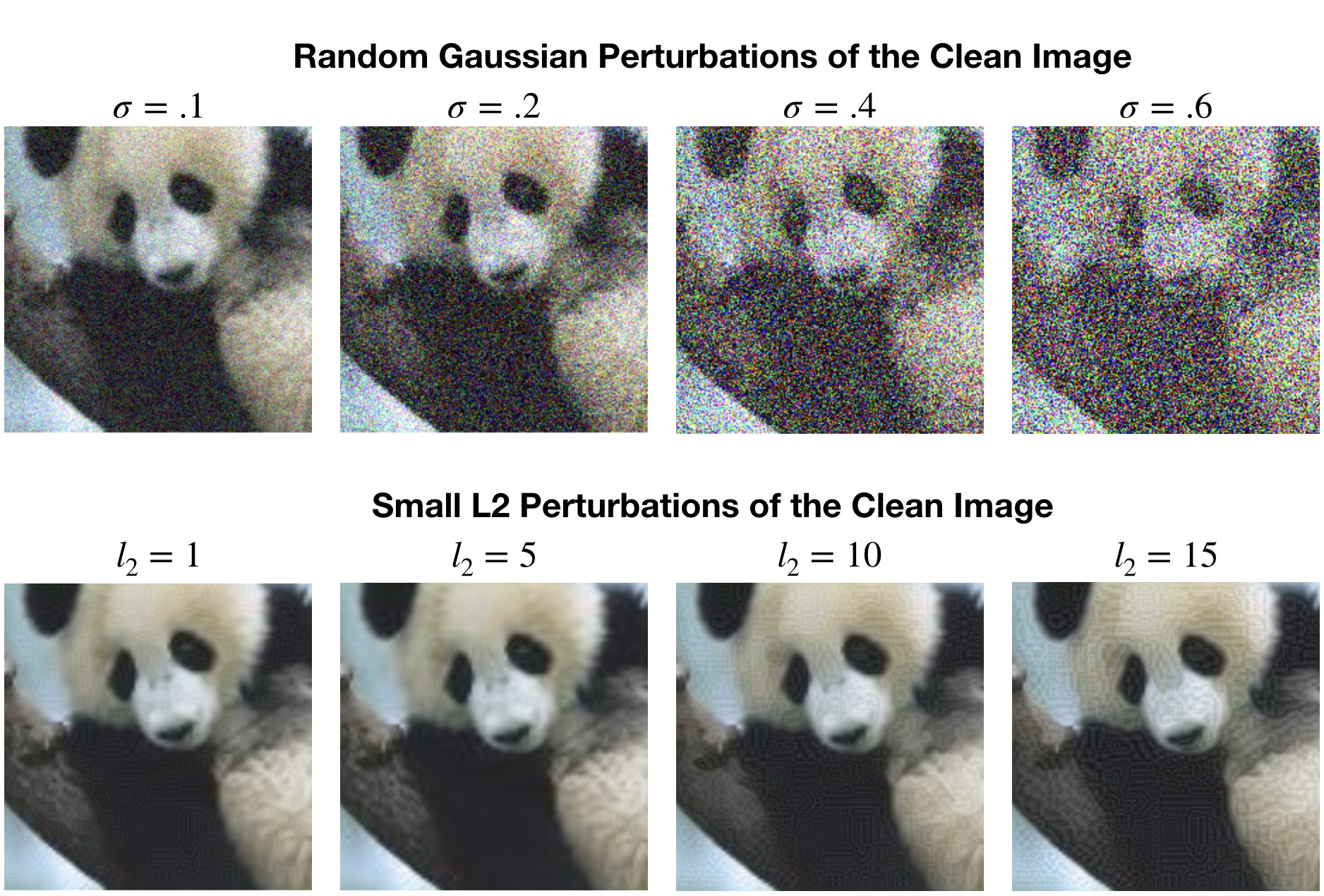}
        \end{subfigure}
        \caption{\textbf{Top:} The optimal curves on ImageNet for different values of $\sigma$. This is both the optimum established by the Gaussian isoperimetric inequality and the relationship described in Equation~\ref{eqn:distscale}. \textbf{Middle:} Visualizing different coordinates of the optimal curves. First, random samples from $x + N(0, \sigma I)$ for different values of $\sigma$. \textbf{Bottom:} Images at different $l_2$ distances from the unperturbed clean image. Each image visualized is the image at the given $l_2$ distance which minimizes visual similarity according to the SSIM metric. Note that images at $l_2 <5$ have almost no perceptible change from the clean image despite the fact that SSIM visual similarity is minimized. }
        \label{fig:optimal_vis}
    \end{figure*}

    In this section we visualize the predicted relationship between worst-case $l_2$ perturbations and generalization in noise as described by Equation~\ref{eqn:distscale} in Section~\ref{sec:cleanimage}. This also visualizes the optimal bound according to the isoperimetric inequality, although the $l_2$ perturbations would be applied to the noisy images themselves rather then clean image. In Figure~\ref{fig:optimal_vis} we plot the optimal curves for various values of $\sigma$, visualize images sampled from $x + N(0, \sigma)$, and visualize images at various $l_2$ distance from the unperturbed clean image. Even for very large noise ($\sigma=.6$), test error needs to be less than $10^{-15}$ in order to have worst-case perturbations be larger than $5.0$. In order to visualize worst-case perturbations at varying $l_2$ distances, we visualize an image that minimizes similarity according to the SSIM metric \citep{MSELoveItOrLeaveIt2009}. These images are found by performing gradient descent to minimize the SSIM metric subject to the containt that $||x - x_{adv}||_2 < \epsilon$.  This illustrates that achieving significant $l_2$ adversarial robustness on Imagenet will likely require obtaining a model that is almost perfectly robust to large Gaussian noise (or a model which significantly violates the linearity assumption from Section~\ref{sec:cleanimage}). To achieve $l_2$ robustness on noisy images, a model \emph{must} be nearly perfect in large Gaussian noise. 

\section{Church Window Plots}
    \label{app:church}
\begin{figure}[h]
        \centering
        \includegraphics[width=0.8\textwidth]{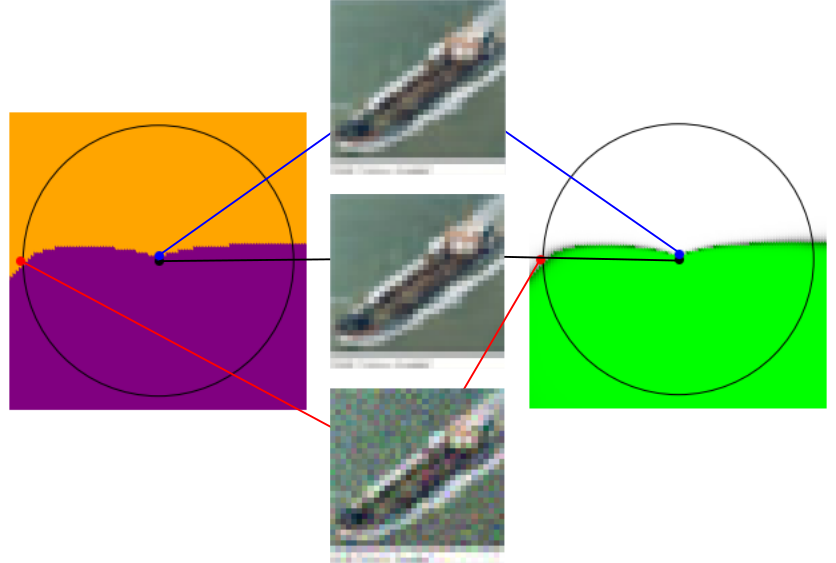}
        \caption{\label{fig:firstchurch}A slice through a clean test point (black, center image), the closest error found using PGD (blue, top image), and a random error found using Gaussian noise (red, bottom image). For this visualization, and all others in this section involving Gaussian noise, we used noise with $\sigma=0.05$, at which the error rate was about 1.7\%. In all of these images, the black circle indicates the distance at which the typical such Gaussian sample will lie. The plot on the right shows the probability that the model assigned to its chosen class. Green indicates a correct prediction, gray or white is an incorrect prediction, and brighter means more confident.}
    \end{figure}

    In figures appearing below, starting at Figure~\ref{fig:firstchurch}, we include many more visualizations of the sorts of church window plots we discussed briefly in Section~4. We will show an ordinarily trained model's predictions on several different slices through the same CIFAR test point which illustrate different aspects of the story told in this paper. These images are best viewed in color.

    \begin{figure}[h]
        \centering
        \includegraphics[width=0.8\textwidth]{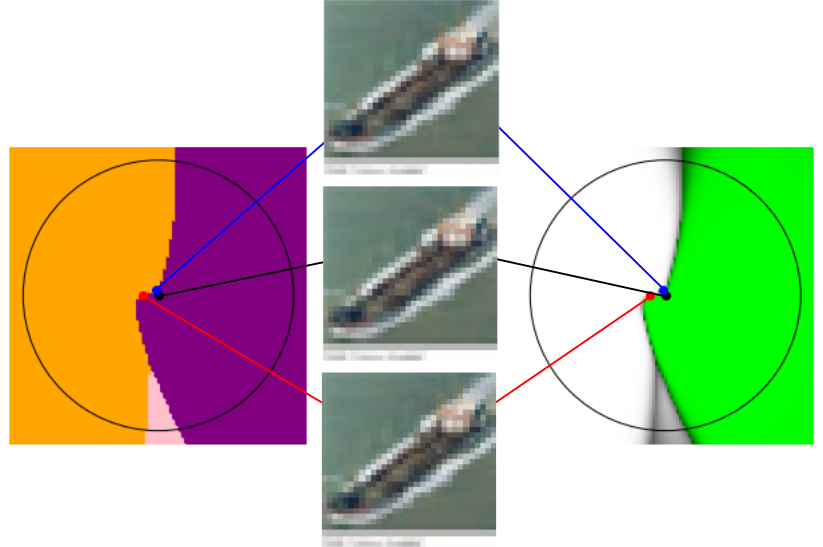}
        \caption{A slice through a clean test point (black, center image), the closest error found using PGD (blue, top image), and the average of a large number of errors randomly found using Gaussian noise (red, bottom image). The distance from the clean image to the PGD error was 0.12, and the distance from the clean image to the averaged error was 0.33. The clean image is assigned the correct class with probability 99.9995\% and the average and PGD errors are assigned the incorrect class with probabilities 55.3\% and 61.4\% respectively. However, it is clear from this image that moving even a small amount into the orange region will increase these latter numbers significantly. For example, the probability assigned to the PGD error can be increased to 99\% by moving it further from the clean image in the same direction by a distance of 0.07.}
    \end{figure}
    \begin{figure}[h]
        \centering
        \includegraphics[width=0.6\textwidth]{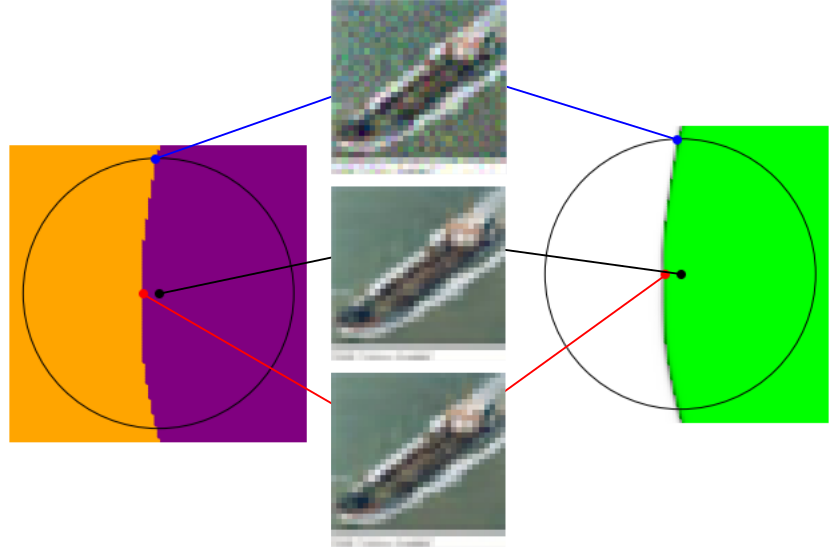}
        \caption{A slice through a clean test point (black, center image), a random error found using Gaussian noise (blue, top image), and the average of a large number of errors randomly found using Gaussian noise (red, bottom image).}
    \end{figure}
    \begin{figure}[h]
        \centering
        \includegraphics[width=0.6\textwidth]{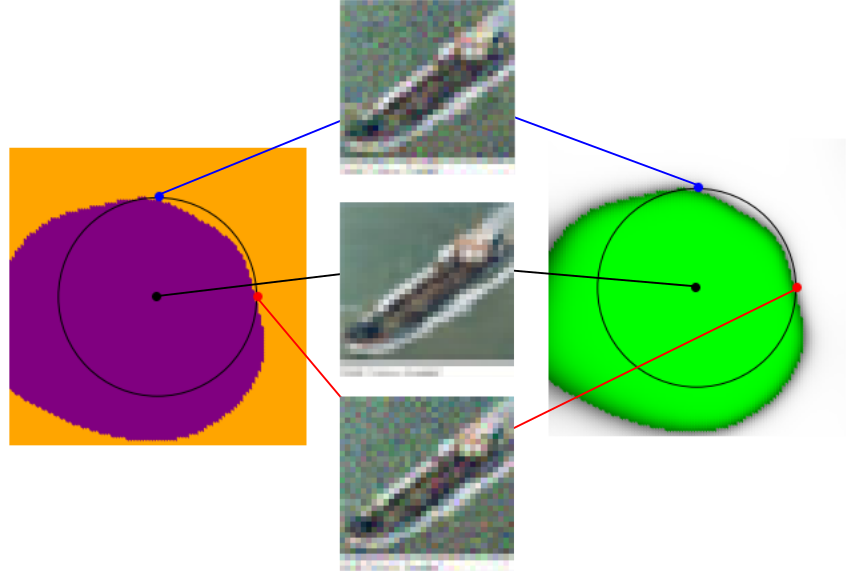}
        \caption{A slice through a clean test point (black, center image) and two random errors found using Gaussian noise (blue and red, top and bottom images). Note that both random errors lie very close to the decision boundary, and in this slice the decision boundary does not appear to come close to the clean image.}
    \end{figure}
    \begin{figure}[h]
        \centering
        \includegraphics[width=0.6\textwidth]{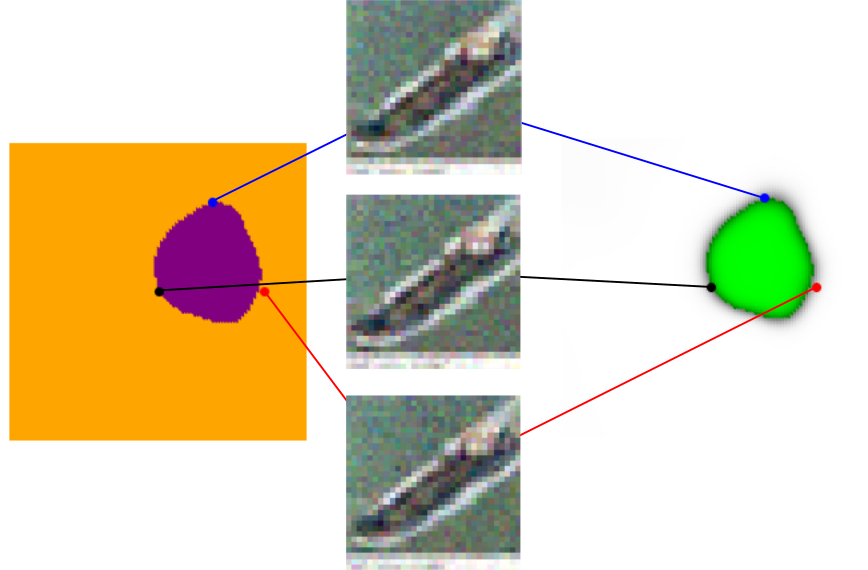}
        \caption{A slice through three random errors found using Gaussian noise. (Note, in particular, that the black point in this visualization does not correspond to the clean image.)}
    \end{figure}
    \begin{figure}[h]
        \centering
        \includegraphics[width=0.65\textwidth]{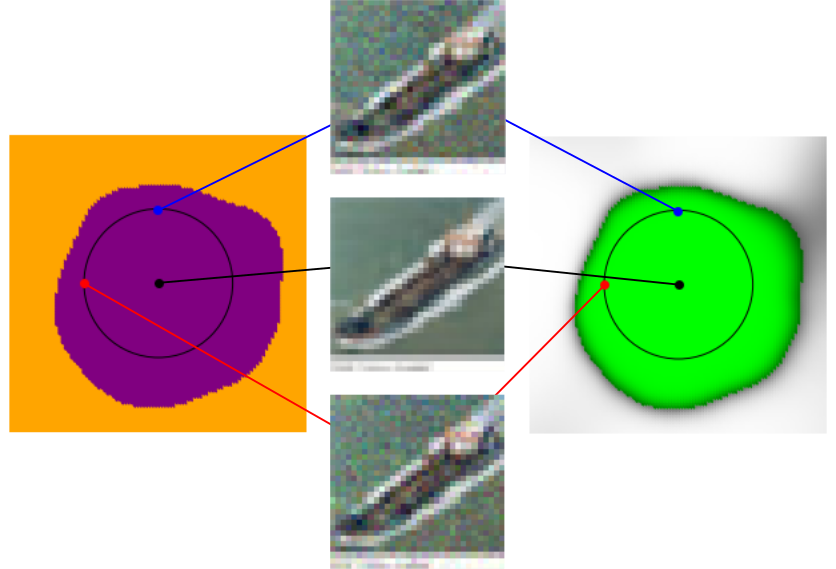}
        \caption{A completely random slice through the clean image.}
    \end{figure}
    \begin{figure}[h]
        \centering
        \begin{subfigure}{0.32\textwidth}
            \includegraphics[width=\textwidth]{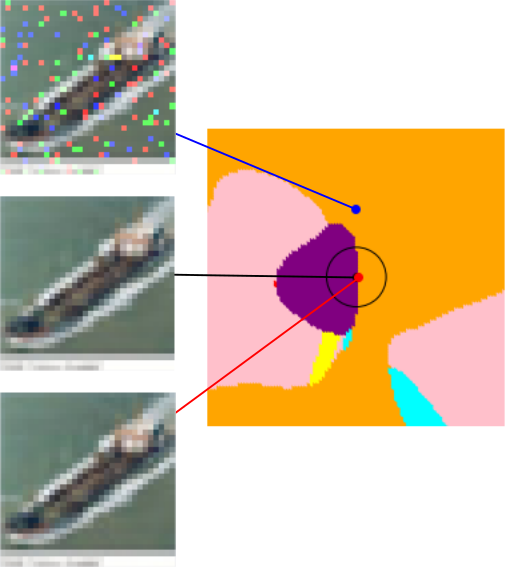}
        \end{subfigure}
        \begin{subfigure}{0.32\textwidth}
            \includegraphics[width=\textwidth]{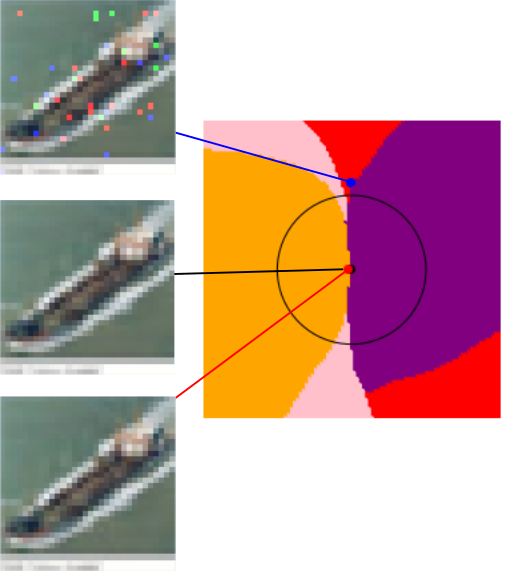}
        \end{subfigure}
        \begin{subfigure}{0.32\textwidth}
            \includegraphics[width=\textwidth]{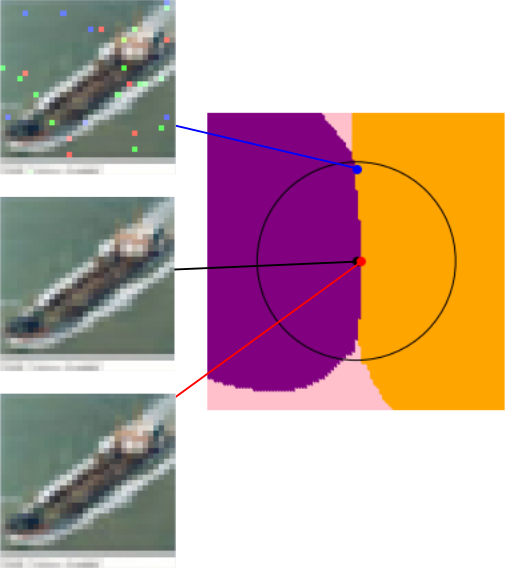}
        \end{subfigure}
        \caption{Some visualizations of the same phenomenon, but using pepper noise rather than Gaussian noise. In all of these visualizations, we see the slice through the clean image (black, center image), the same PGD error as above (red, bottom image), and a random error found using pepper noise (blue, top image). In the visualization on the left, we used an amount of noise that places the noisy image further from the clean image than in the Gaussian cases we considered above. In the visualization in the center, we selected a noisy image which was assigned to neither the correct class nor the class of the PGD error. In the visualization on the right, we selected a noisy image which was assigned to the same class as the PGD error.}
    \end{figure}
    \begin{figure}[h]
        \centering
        \includegraphics[width=0.8\linewidth]{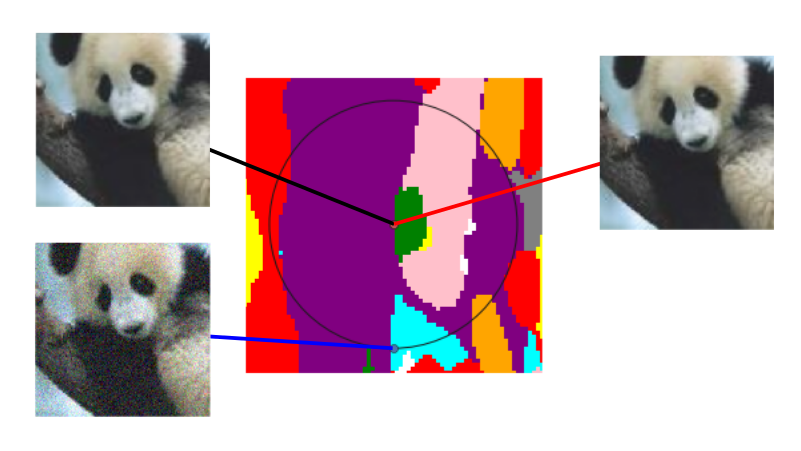}
        \caption{Not all slices containing a PGD error and a random error look like Figure~\ref{fig:pandas}. This image shows a different PGD error which is assigned to a different class than the random error.}
    \end{figure}
    \begin{figure}[h]
        \centering
        \includegraphics[width=0.8\textwidth]{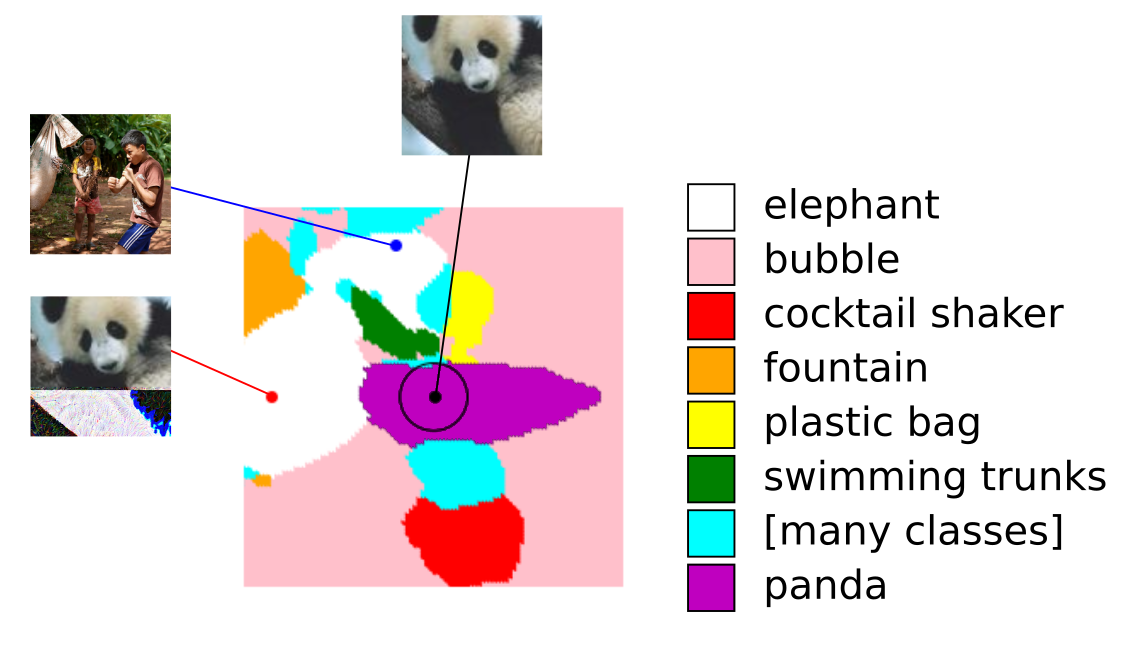}
        \caption{A slice with the same black point as in Figure~3 from the main text, together with an error from the clean set (blue) and an adversarially constructed error (red) which are both assigned to the same class (``elephant''). We see a different slice through the same test point but at a larger scale. This slice includes an ordinary test error along with an adversarial perturbation of the center image constructed with the goal of maintaining visual similarity while having a large $l_2$ distance. The two errors are both classified (incorrectly) by the model as ``elephant.'' This adversarial error is actually \emph{farther} from the center than the test error, but they still clearly belong to the same connected component. This suggests that defending against worst-case content-preserving perturbations \citep{gilmer2018motivating} requires removing all errors at a scale comparable to the distance between unrelated pairs of images.}
    \end{figure}

\section{The Distribution of Error Rates in Noise}
    \label{app:noisehist}
\begin{figure}[h]
        \centering
        \includegraphics[width=0.65\textwidth]{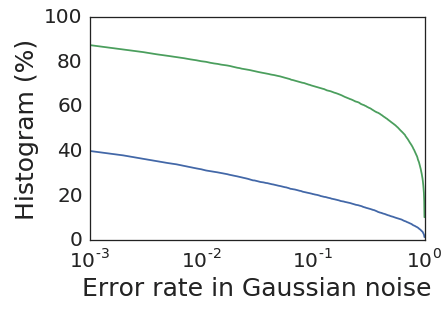}
        \caption{\label{fig:noisehist}The cdf of the error rates in noise for images in the test set. The blue curve corresponds to a model trained and tested on noise with $\sigma=0.1$, and the green curve is for a model trained and tested at $\sigma=0.3$. For example, the left most point on the blue curve indicates that about 40\% of test images had an error rate of at least $10^{-3}$.}
    \end{figure}

    Using some of the models that were trained on noise, we computed, for each image in the CIFAR test set, the probably that a random Gaussian perturbation will be misclassified. A histogram is shown in Figure~\ref{fig:noisehist}. Note that, even though these models were trained on noise, there are still many errors around most images in the test set. While it would have been possible for the reduced performance in noise to be due to only a few test points, we see clearly that this is not the case.

\section{A Collection of Model Errors}

\begin{figure}[h]
      \centering
      \includegraphics[width=1.0\linewidth]{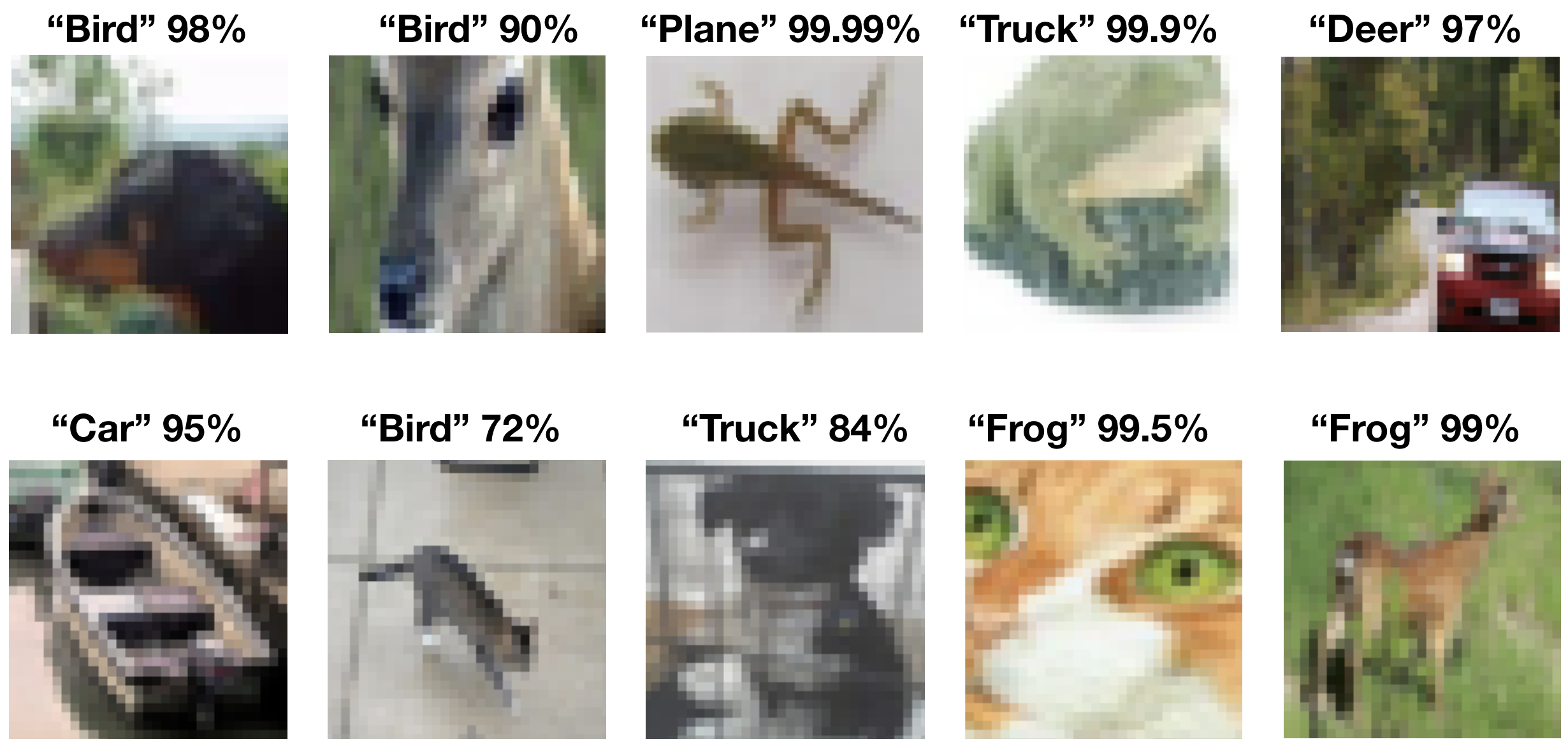}
      \caption{\label{fig:firsterror}A collection of adversarially chosen model errors. These errors appeared in the ImageNet validation set. Despite the high accuracy of the model there remain plenty of errors in the test set that a human would not make.}
    \end{figure}

    Finally, in the figures starting at Figure~\ref{fig:firsterror} we first show a collection of iid test errors for the ResNet-50 model on the ImageNet validation set. We also visualize the severity of the different noise distributions considered in this work, along with model errors found by random sampling in these distributions. 
    
    \label{app:nonhuman_cifar}
    \label{app:nonhuman}
    \begin{figure}[h]
      \centering
      \includegraphics[width=1.0\linewidth]{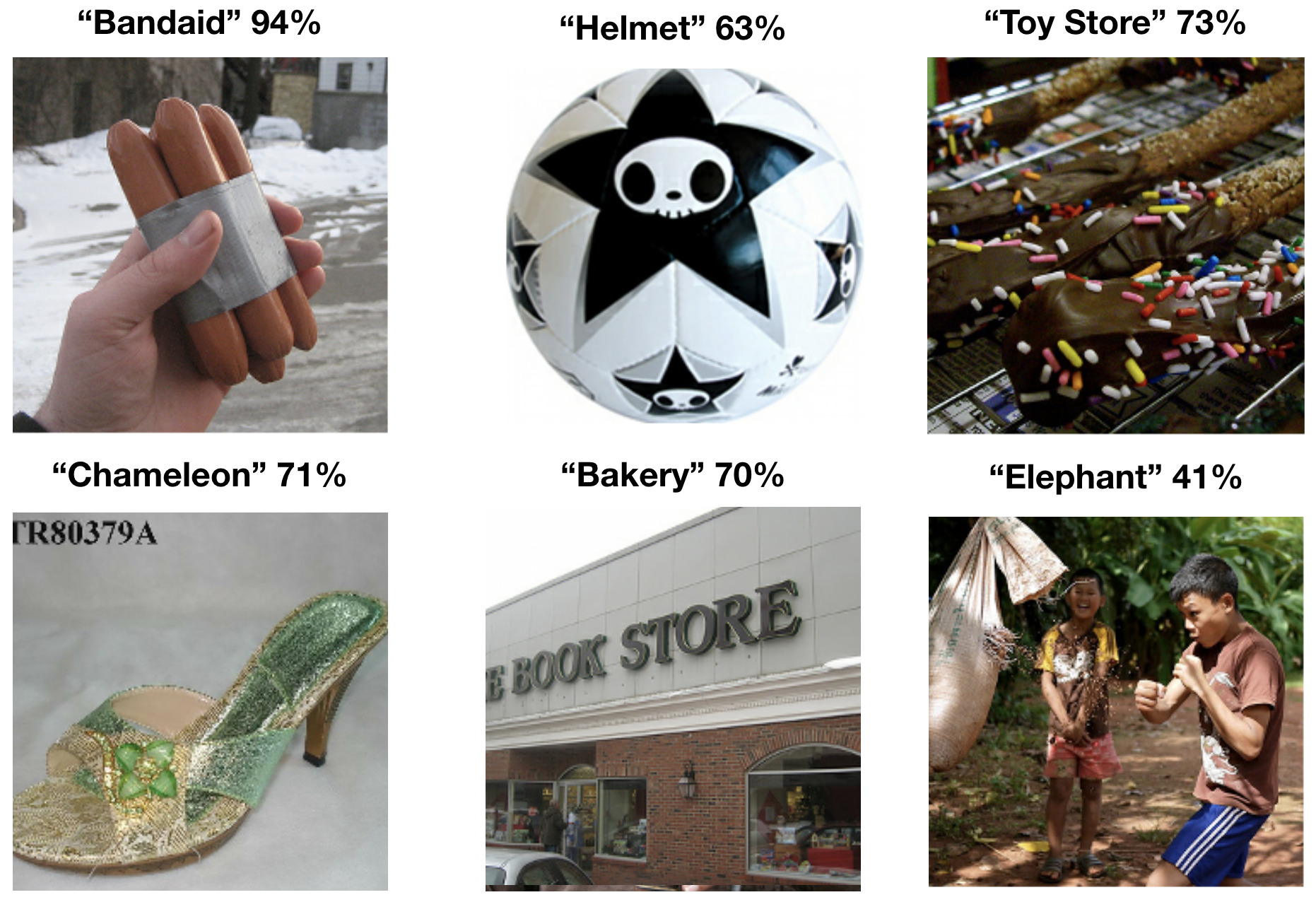}
      \caption{A collection of adversarially chosen model errors. These errors appeared in the ImageNet validation set. Despite the high accuracy of the model there remain plenty of errors in the test set that a human would not make.}
    \end{figure}

    \begin{figure}[h]
      \centering
      \includegraphics[width=1.0\linewidth]{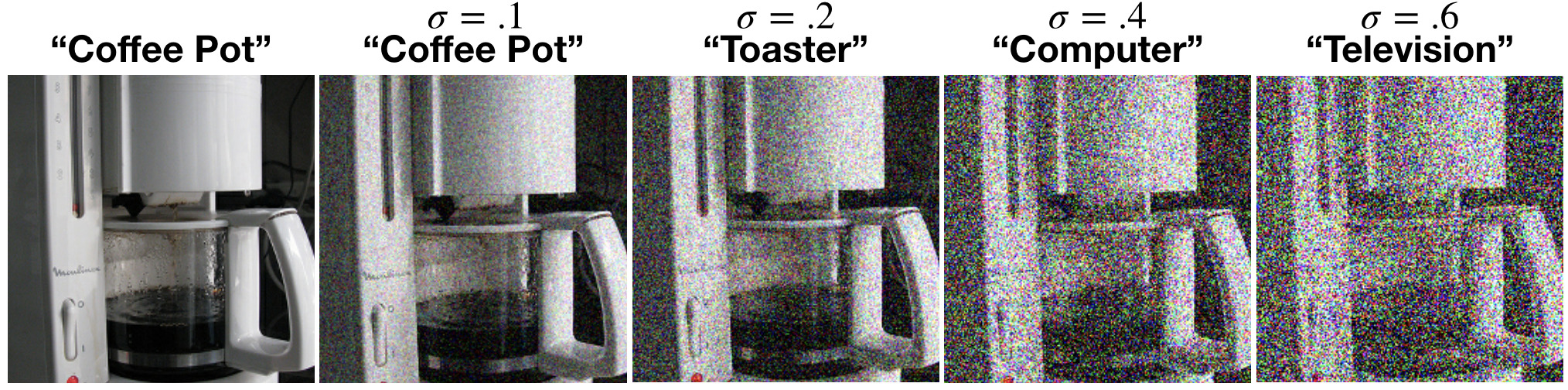}
      \caption{Visualizing the severity of Gaussian noise, along with model errors found in this noise distribution. Note the model shown here was trained at noise level $\sigma=.6$.}
    \end{figure}

\end{document}